%% file: OJSP.tex
\documentclass[journal]{IEEEtran}
\usepackage[dvips]{graphicx}
\usepackage{algorithm}
\usepackage{algorithmic}
\usepackage{psfrag}
\usepackage{bm}
\usepackage{amssymb,mathtools}
\usepackage{amsfonts}
\usepackage{amsmath,amssymb}
\usepackage{amsthm}
\usepackage{cite} 
\usepackage[dvipsnames]{xcolor}
\usepackage{url}
\usepackage[letterpaper, left=0.625in, right=0.625in, bottom=1in, top=0.75in]{geometry}
\usepackage[hyperfootnotes=false,hidelinks]{hyperref}
\usepackage{graphicx}
\usepackage{upgreek}
\usepackage{yfonts}
\usepackage{cleveref}
\usepackage{amsmath}
\usepackage{arydshln}
\usepackage{soul}
\usepackage{mathtools}
\usepackage{balance}
\input{symbols.tex}

\title{Noise-Robust and Resource-Efficient ADMM-based Federated Learning} 

\author{Ehsan Lari, \IEEEmembership{Student Member, IEEE}, Reza Arablouei, Vinay Chakravarthi Gogineni, \IEEEmembership{Senior Member, IEEE}, and Stefan Werner, \IEEEmembership{Fellow, IEEE}
\thanks{This work was supported by the Research Council of Norway. Parts of this work have been presented at the 2023 IEEE Statistical Signal Processing Workshop (SSP) [DOI: 10.1109/SSP53291.2023.10208024] and the 2023 Asia Pacific Signal and Information Processing Association Annual Summit and Conference (APSIPA ASC) [DOI: 10.1109/APSIPAASC58517.2023.10317446]. \textit{(Corresponding author: Ehsan Lari.)}}
\thanks{Ehsan Lari and Stefan Werner are with the Department of Electronic Systems, Norwegian University of Science and Technology, 7491 Trondheim, Norway (e-mail: \{ehsan.lari, stefan.werner\}@ntnu.no). Stefan Werner is also with the Department of Information and Communications Engineering, Aalto University, 00076, Finland.}
\thanks{Reza Arablouei is with the Commonwealth Scientific and Industrial Research Organisation, Pullenvale, QLD 4069, Australia (e-mail: reza.arablouei@csiro.au).}
\thanks{Vinay Chakravarthi Gogineni is with the SDU Applied AI and Data Science, The Maersk Mc-Kinney Moller Institute, Faculty of Engineering, University of Southern Denmark, Odense 5230, Denmark (e-mail: vigo@mmmi.sdu.dk).}}

\begin{document}

\makeatother

\newcommand{\Lim}[1]{\raisebox{0.5ex}{\scalebox{0.8}{$\displaystyle \lim_{#1}\;$}}}

\maketitle

\begin{abstract}

Federated learning (FL) leverages client-server communications to train global models on decentralized data. However, communication noise or errors can impair model accuracy. To address this problem, we propose a novel FL algorithm that enhances robustness against communication noise while also reducing communication load. We derive the proposed algorithm through solving the weighted least-squares (WLS) regression problem as an illustrative example. We first frame WLS regression as a distributed convex optimization problem over a federated network employing random scheduling for improved communication efficiency. We then apply the alternating direction method of multipliers (ADMM) to iteratively solve this problem. To counteract the detrimental effects of cumulative communication noise, we introduce a key modification by eliminating the dual variable and implementing a new local model update at each participating client. This subtle yet effective change results in using a single noisy global model update at each client instead of two, improving robustness against additive communication noise. Furthermore, we incorporate another modification enabling clients to continue local updates even when not selected by the server, leading to substantial performance improvements. Our theoretical analysis confirms the convergence of our algorithm in both mean and the mean-square senses, even when the server communicates with a random subset of clients over noisy links at each iteration. Numerical results validate the effectiveness of our proposed algorithm and corroborate our theoretical findings.

\end{abstract}


\section{Introduction}

\IEEEPARstart{T}{he}
proliferation of smart devices has led to the widespread availability of big data, which has the potential to enhance decision-making processes for end-users~\cite{mcmahan2017communication, smith2017federated, 10153432, 9062302, lari2024privacy}. However, leveraging this data effectively presents several challenges, particularly in terms of privacy, security, and resource management. A key issue is that data is typically stored locally on edge devices, and transferring it to a central server or the cloud raises privacy and security concerns and can lead to excessive resource utilization. Federated learning (FL) has emerged as a promising machine learning paradigm that addresses these challenges by enabling edge devices to collaboratively train a global model without sharing their locally stored data~\cite{mcmahan2017communication,9170559}. FL faces two significant challenges: data heterogeneity and device heterogeneity~\cite{yang2019federated}. Data heterogeneity arises from the non-independent and identically distributed (non-IID) nature of data or imbalances among client datasets used in training the global model~\cite{9528995,9084352}. Diversity in data distribution across participants can significantly impact model convergence and performance. Additionally, device heterogeneity refers to the disparities in storage capacity, energy resources, computational power, and communication capabilities among participating clients~\cite{9060868,9770266}. Variations in device characteristics can affect the efficiency and fairness of the FL process. Despite these challenges, FL emerges as a promising approach for managing heterogeneous data and devices in distributed learning settings.

The FL literature frequently employs the FedAvg algorithm~\cite{mcmahan2017communication} as a standard benchmark. FedAvg begins with the server broadcasting its aggregated global model to a randomly chosen subset of clients, typically over a wireless network. These clients then perform local training to refine their local models before sending the updates back to the server. The server aggregates these local models into a new global model, repeating this process until a specific convergence criterion is met. Building on FedAvg, numerous FL approaches have been developed to address various challenges inherent in distributed learning. These challenges include preserving privacy~\cite{10420449, 10018276, 10064312, 9308910, 9055478, 9069945, 10290905}, mitigating Byzantine attacks~\cite{lari2024analyzing, icasspehsan, fung2018mitigating, 10095468}, and improving communication efficiency~\cite{10429818, 9878377, 10095356, vargaftik2022eden, 9933811, 9746228}. However, many of these approaches assume ideal communication links, overlooking the potential impact of communication errors or noise~\cite{9751160,8664630,8647927,8761315}. In practical deployments, the communication channels between clients and the server are often subject to noise, which can degrade model performance~\cite{9026922}. To address this issue, various techniques have been proposed to enhance the robustness of FL models in the presence of communication noise. While some studies focus on the uplink noise and neglect the downlink noise \cite{9014530,9119095}, others account for noise in both communication directions~\cite{9515709}. 

In real-world scenarios, communication channels between the central server and edge devices are often imperfect, leading to noise in both uplink and downlink~\cite{zhou2021communication,9026922,9277666}. This leads to the server receiving noisy versions of local model updates and clients obtaining noisy versions of the aggregated model from the server, potentially compromising the quality of the learned model. The presence of noise in communication channels can severely impair the performance of gradient-descent-based FL algorithms. Several solutions have been proposed to address this challenge. One approach involves utilizing a new loss function that incorporates the first-order derivative as a regularizer to counteract additive noise~\cite{9026922}. Additionally, resilience to downlink noise can be enhanced through techniques such as digital transmission with quantization and channel coding or analog downlink transmission using uncoded global model updates~\cite{9515709}. Moreover, \cite{9500833,9672092} propose that controlling the scale of the communication signal-to-noise ratio (SNR) can help tolerate noise while maintaining the convergence rate of FedAvg. 
In \cite{9322580}, the authors utilize precoding and scaling during transmissions to mitigate the adverse effects of noisy channels, ensuring the convergence of their algorithm. While these methods show promise, they typically require additional resources on the client side. This requirement can pose challenges in practical FL applications, where clients may operate under constraints such as limited power, energy, memory, or computational capacity.

In \cite{4407653, 4982756, lari2024distributed}, it is shown that the algorithms based on the alternating direction method of multipliers (ADMM) \cite{boydadmm} can achieve convergence even in the presence of additive communication noise. This robustness positions ADMM as a promising candidate for FL applications where communication noise is a concern. However, conventional ADMM algorithms require the participation of all clients in each FL round, which may be impractical in real-world FL scenarios due to resource constraints and the heterogeneity of edge devices. Therefore, there is a pressing need for developing ADMM-based FL algorithms that can effectively cope with communication noise with minimal additional communication or computational overhead.

In this paper, we propose a new communication-efficient FL algorithm based on ADMM that is resilient against communication noise and errors in both uplink and downlink channels, without imposing any additional computational burden on participating clients. Using the weighted least-squares (WLS) regression problem as a motivating example, we develop our proposed FL algorithm by iteratively solving an aptly formulated distributed convex optimization problem via ADMM. To enhance communication efficiency, we incorporate random client scheduling, where the server selects a subset of clients during each FL iteration. To alleviate the adverse effects of link noise, we communicate a linear combination of the last two global model updates and eliminate the dual model parameters at all participating edge devices. Moreover, we enable clients to continue performing model updates, even when not scheduled for participation in any FL round. We evaluate the proposed algorithm and showcase its effectiveness through comprehensive theoretical performance analysis and numerical simulations. 
In summary, our main contributions are:
\begin{itemize}
    \item We propose an ADMM-based FL algorithm that is robust to communication noise without imposing any additional computational overhead on clients.
    \item We employ a random client scheduling mechanism that effectively reduces the communication load during the learning process, distinguishing our approach from conventional ADMM algorithms.T In addition, to further enhance performance, we allow continual local updates at clients not selected by the server in any FL round.
    \item As a key advancement over the related conference precursors~\cite{sspehsan,apsipaehsan}, we present a rigorous first- and second-order theoretical analysis of the proposed algorithm, proving its convergence in both mean and mean-square senses, even in the presence of link noise and random scheduling. In addition, we derive the theoretical steady-state mean-square error (MSE) as a function of various parameters, including the number of participating clients in each iteration and the uplink and downlink noise variances. We validate the accuracy of our analysis through comparisons with simulation results.
\end{itemize}

The remainder of this paper is organized as follows. In section \ref{sec:pre}, we provide an overview of the system model and the background for our problem. In section \ref{sec:rerce}, we introduce our proposed noise-robust and communication-efficient FL algorithm. In addition, we introduce a modification to allow continual local updates at clients regardless of random client scheduling to improve performance. In section \ref{sec:analyis}, we evaluate the robustness of the proposed algorithm by analyzing its theoretical mean and mean-square convergence and calculating its theoretical steady-state MSE. In section \ref{sec:simulation}, we validate our theoretical findings through comprehensive numerical experiments, assessing algorithm performance through both theoretical predictions and numerical simulations. Finally, in section \ref{sec:conc}, we present some concluding remarks.

\section{Federated Learning over Noisy Channels} \label{sec:pre}

We consider a federated network consisting of a server and $K$ clients, where the clients communicate with the server over wireless channels. Each client $k$ has access to a local dataset denoted by $\Dcal_{k}=\{\Xbf_k, \ybf_k\}$, which comprises a column response vector $\ybf_k$ with $d_{k}$ entries, and a data matrix $\Xbf_k$ of size $d_{k} \times L$. For each client $k$, a linear model is employed to relate the data matrix $\Xbf_k$ to the response vector $\ybf_k$ as
\begin{equation}
   \ybf_k = \Xbf_k \boldsymbol\omega + \boldsymbol{\nu}_{k},
\label{eq:ModelFL}
\end{equation}
where $\boldsymbol\omega$ denotes the global regression model parameter vector of size $L$, and $\boldsymbol{\nu}_{k}$ represents the observation noise or perturbation, which is a vector of size $d_{k}$ with each entry assumed to be zero-mean Gaussian.

\subsection{Federated Weighted Least-Squares Regression}

Weighted least-squares (WLS) regression is a natural extension of least-squares regression that offers significant advantages in various signal processing applications, such as power system state estimation \cite{6279588}, position estimation \cite{5976479}, and image noise reduction \cite{farbman2008edge}. In WLS, different observations are assigned weights based on their reliability, allowing the model to more accurately reflect the data when observations vary in quality. This approach can effectively mitigate the impact of less reliable data, resulting in more precise and dependable models. 

In federated WLS regression, the goal is to collaboratively estimate the global model parameter vector $\boldsymbol\omega$ by minimizing a global objective function across a federated network. This is framed as a global WLS estimation problem within the FL framework as
\begin{align} \label{eq:Model3}
\min_{\{ \wbf_k \}} & \sum\limits_{k=1}^{K} \Jcal_k(\wbf_k) \notag \\
\mathrm{s.t.} & \ \wbf_k = \wbf, \ k \in \{ 1, 2, \cdots, K \},
\end{align}
where $\Jcal_k(\wbf_k) = \| \ybf_k - \Xbf_k \wbf_k \|_{\Wcalbf_k}^2$ is the local objective function for estimating $\boldsymbol\omega$ at client $k$, $\Wcalbf_k$ is the weight matrix specific to client $k$, $\wbf_k$ is the local model estimate at client $k$, and $\wbf$ serves as the global model estimate.
The optimal solution to \eqref{eq:Model3}, which can be viewed as a federated version of distributed Pareto optimization~\cite{a17337, 6461050} for WLS regression, is stated as
\begin{equation} \label{eq:OptSol}
    \wbf^{\star} = \left( \sum_{k=1}^{K} \Xbf_k^{\intercal} \Wcalbf_k \Xbf_k \right)^{-1} \left( \sum_{k=1}^{K} \Xbf_k^{\intercal} \Wcalbf_k \ybf_k \right).
\end{equation}

To solve \eqref{eq:Model3} within an FL framework, we utilize the ADMM algorithm \cite{boydadmm}. Therefore, we express the augmented Lagrangian function corresponding to \eqref{eq:Model3} as
\begin{align} \label{eq:FD1}
   \Lcal (\wbf_k,\wbf,\zbf_k) & = \sum\limits_{k=1}^{K} \Lcal_k(\wbf_k,\wbf,\zbf_k) \\
   & = \sum\limits_{k=1}^{K}  \Jcal_k( \wbf_k ) + \langle \wbf_k - \wbf,\zbf_k \rangle + \frac{\rho}{2} \| \wbf_k - \wbf \|_2^2, \notag 
\end{align}
where $\zbf_k$ is the Lagrange multiplier vector and and $\rho > 0$ is the penalty parameter. Consequently, we derive the recursive update equations at each client $k$ and iteration number $n$ as
\begin{subequations}\label{eq:FD}
\begin{align}
 \zbf_{k,n} & = \zbf_{k,n-1} + \rho (\wbf_{k,n} - {\wbf}_k) \label{eq:FDpi1}
\\ \wbf_{k,n+1} & = \widehat{\wbf}_k - \Nbf_k ( \zbf_{k,n} - \rho  {\wbf}_k), \label{eq:FDx1}
\end{align}
\end{subequations}
along with the recursive update equation at the server as
\begin{equation} \label{eq:FDxbar1}
    \wbf_{n+1} =  \frac{1}{K} \sum\limits_{k=1}^{K} \left( {\wbf}_{k,n+1} + \frac{1}{\rho}{\zbf}_{k,n} \right),
\end{equation}
where we define 
\begin{equation}
    \Nbf_k = \left( 2 \Xbf_k^{\intercal} \Wcalbf_k \Xbf_k + \rho \Ibf \right)^{-1}
\end{equation}
and 
\begin{equation}\label{w0}
    \widehat{\wbf}_k = 2 \Nbf_k \Xbf_k^{\intercal}  \Wcalbf_k \ybf_k.
\end{equation}

In this solution, after performing local training, i.e., \eqref{eq:FDpi1} and \eqref{eq:FDx1}, each client shares its local estimate of $ \wbf_{k,n+1} + \rho^{-1} \zbf_{k,n}$ with the server. The server then obtains the global estimate as in~\eqref{eq:FDxbar1} and broadcasts it to all clients while the FL process continues.

\subsection{Dual Variable Elimination}

We posit that, in the recursions~\eqref{eq:FD}-\eqref{eq:FDxbar1}, it is necessary to transmit a combination of the primal and dual model parameter estimates to the server to enable the aggregation that produces the global model estimate. However, the dual update information can be integrated into the primal update by judiciously selecting the initial estimates and introducing a new local primal update at clients.
Accordingly, we reformulate \eqref{eq:FD}-\eqref{eq:FDxbar1} as
\begin{subequations}\label{fl2}
\begin{align}
\wbf_{n} &= \frac{1}{K} \sum\limits_{k=1}^{K} {\wbf}_{k,n} \label{9a}\\
\wbf_{k,n+1} &= ( \Ibf - \rho \Nbf_k)\wbf_{k,n}+\rho \Nbf_k(2\wbf_n-\wbf_{n-1}) \label{9b}
\end{align}
\end{subequations}
by initializing with $\wbf_{-1}=\mathbf{0}$, $\zbf_{k,-1}=\mathbf{0}$, and $\wbf_{k,0}=\widehat{\wbf}_k$, hence eliminating the Lagrange multipliers $\zbf_{k,n}$. The recursion begins with clients sharing their $\widehat{\wbf}_k$ with the server, which then aggregates them and broadcasts the resulting global model estimate to the clients. 
We further define $\sbf_n=2\wbf_n-\wbf_{n-1}$ and modify~\eqref{9b} as
\begin{align}
\wbf_{k,n+1} &= ( \Ibf-\rho \Nbf_k)\wbf_{k,n}+\rho \Nbf_k\sbf_n. \label{fl3} 
\end{align}
A key advantage of \eqref{fl3} over \eqref{9b} is that the linear combination of the two most recent global model estimates, required for local model updates, is performed at the server rather than by individual clients. This can significantly reduce the impact of communication noise as the combination occurs before transmission to clients.

\subsection{Communication Noise}

Clients and the server often communicate via wireless channels, where both uplink and downlink channels are susceptible to noise. In the downlink, clients receive noisy versions of the aggregated model updates from the server. Specifically, at iteration $n$, client $k$ receives $\tilde{\sbf}_{k,n} = \sbf_n + \boldsymbol{\zeta}_{k,n}$ where $\boldsymbol{\zeta}_{k,n}$ represents the downlink noise affecting the transmission
In the uplink, the server receives a noisy version of each client's local model update, i.e., 
$\tilde{\wbf}_{k,n+1} = \wbf_{k,n+1} + \boldsymbol{\eta}_{k,n}$
where $\boldsymbol{\eta}_{k,n}$ denotes the uplink noise for client $k$ at iteration $n$. Considering the impact of communication noise and allowing client updates to occur before server aggregation, we obtain
\begin{subequations}\label{fln}
\begin{align}
\wbf_{k,n+1} &= ( \Ibf - \rho \Nbf_k)\wbf_{k,n}+\rho \Nbf_k\tilde{\sbf}_{k,n} \label{flnb}\\
\wbf_{n+1} &= \frac{1}{K} \sum\limits_{k=1}^{K} {\tilde\wbf}_{k,n+1} \label{flna}
\end{align}
\end{subequations}

\section{Resource-efficient FL over Noisy Channels} \label{sec:rerce}

In this section, we introduce RERCE-Fed, our proposed FL algorithm, that is both resource-efficient and robust to communication errors.

\subsection{Random Scheduling}\label{rs}

In \eqref{fln}, there is an implicit assumption that all clients participate in each global model update iteration. However, in FL, clients often have limited communication and computational resources. Therefore, requiring participation of all clients in every global update round can lead to significant drawbacks, such as prolonged convergence time or excessive resource utilization. To address this challenge, we enable the server to implement a mechanism known as random scheduling. This approach involves randomly selecting a subset of clients, denoted by $\Scal_n$, to participate in model aggregation at each iteration $n$. We consider the cardinality of this subset, $\Ccal = |\Scal_n|$, to be fixed throughout the FL process.

\subsection{RERCE-Fed}

In our proposed algorithm, during each global iteration $n$, the selected clients, $k \in \Scal_n$, receive $\tilde{\sbf}_{k,n}$ from the server and update their models. The server then receives $\tilde{\wbf}_{k,n+1}$ from these clients and aggregates them. Subsequently, it broadcasts the most recent global update to a new set of selected clients in the next iteration. Clients not selected by the server in a given round retain their latest local update until they are selected again. Therefore, the recursions of the proposed RERCE-Fed algorithm are expressed as \cite{sspehsan}
\begin{subequations} \label{eq:rercefed}
\begin{align}
\wbf_{k,n+1} & = \left(\Ibf-a_{k,n}\rho\Nbf_k\right)\wbf_{k,n}+ a_{k,n} \rho  \Nbf_k  \tilde{\sbf}_{k,n}
\label{eq:FDx3} \\
\wbf_{n+1} & = \frac{1}{ \Ccal } \sum\limits_{k=1}^{K} a_{k,n} \tilde{\wbf}_{k,n+1}, \label{eq:FDxbar4}
\end{align}
\end{subequations}
where $a_{k,n}$ is the indicator variable for random scheduling, with $a_{k,n} = 1$ when client $k$ is selected by the server in iteration $n$ (i.e., $k \in \Scal_n$) and $a_{k,n} = 0$ otherwise. We summarize RERCE-Fed in Algorithm \ref{Alg:RERCE_Fed}. 

\begin{algorithm}[t!]
\textbf{Parameters}: penalty parameter $\rho$, number of clients $K$, set of clients $ \Scal$ \\
\textbf{Initialization}: global model $ \wbf_0 =  \wbf_{-1} = \mathbf{0}$, local models $\wbf_{k,0} = \widehat{\wbf}_k$\\[1mm]
\textbf{For} \,$n=1,\cdots,$ \textit{Until Convergence} \\[1mm]
\hspace*{5mm} The server randomly selects a subset $\Scal_n$ of its clients and sends the aggregated global model $\sbf_n$ to them. \\[1mm]
\textbf{Client Local Update}:\\[1mm]
\hspace*{5mm} \textbf{If} $k \in \Scal_n$ \\[1mm]
\hspace*{10mm} Receive $\tilde{\sbf}_{k,n}$, a noisy version of $\sbf_{n}$, from the server. \\
\hspace*{10mm} Update the local model as\\[1mm] 
\hspace*{15mm} $\wbf_{k,n+1} = ( \Ibf - \rho \Nbf_k ) \wbf_{k,n}+ \rho \Nbf_k \tilde{\sbf}_{k,n}$ \\[1mm]
\hspace*{10mm} Send $\wbf_{k,n+1}$ to the server. \\[1mm]
\hspace*{5mm}\textbf{EndIf} \\[1mm]
\textbf{Aggregation at the Server}:\\[1mm]
\hspace*{5mm} The server receives $\tilde{\wbf}_{k,n+1}$, noisy versions of the locally updated models from the selected clients $k \in \Scal_n$ and aggregates them via\\[1mm]
\hspace*{5mm} $\wbf_{n+1} = \frac{1}{ \Ccal } \sum\limits_{k \in \Scal_n} \tilde{\wbf}_{k,n+1}$\\
\hspace*{6.4mm} $\sbf_{n+1} = 2 \wbf_{n+1} - \wbf_{n}$\\[1mm]
\textbf{EndFor}\\
\caption{RERCE-Fed.} \label{Alg:RERCE_Fed}
\end{algorithm} 

\subsection{RERCE-Fed with Continual Local Updates} \label{sec:continual}

Unlike the conventional random scheduling approach, described in section~\ref{rs}, where non-selected clients refrain from local updates and their latest model estimates are not integrated into the global aggregation process, we propose a new approach where all clients, regardless of their selection status, continually update their local model estimates during each iteration. 
This new approach can enhance overall performance without introducing any additional communication overhead or imposing any notable increase in computational load of clients or the server, as we will demonstrate later. To implement this approach, clients store the most recent global model estimate received from the server, while the server holds the latest local model estimates received from all clients. Consequently, clients continually update their local models using the most recent global model estimate, and the server updates the global model using the latest local updates from all clients, irrespective of random scheduling. When a client is selected at iteration $n$, its latest local model estimate is synchronized at the server, and the global model estimate received from the server supersedes the previous version at the client. 

Therefore, the recursions of the RERCE-Fed algorithm with continual local updates are given by \cite{apsipaehsan}
\begin{subequations} \label{eq:clu0}
\begin{align}
\wbf_{k,n+1} &=( \Ibf-\rho \Nbf_{k})\wbf_{k,n} \notag\\
&\quad +\rho \Nbf_{k}\left[a_{k,n}\tilde{\sbf}_{k,n}+(1-a_{k,n})\tilde{\sbf}_{k,m}\right]\label{eq:FDx3_20}\\
\wbf_{n+1} &=\frac{1}{K}\sum\limits_{k=1}^{K} \left[a_{k,n}\tilde{\wbf}_{k,n+1}+(1-a_{k,n})\tilde{\wbf}_{k,m}\right],\label{eq:FDxbar4_20}
\end{align}
\end{subequations}
where $\tilde{\sbf}_{k,m}$ denotes the most recent global model estimate received from the server and stored in client $k$. This estimate is utilized when the client is not selected by the server. Additionally, $\tilde{\wbf}_{k,m}$ represents the most recent local model estimate associated with client $k$, which is stored at the server and utilized during iterations when the client is not chosen through random scheduling. Defining $\tbf_{k,n+1} = 2\wbf_{k,n+1}-\wbf_{k,n}$, we can restate \eqref{eq:FDxbar4_20} as
\begin{align} 
\sbf_{n+1} &=\frac{1}{K}\sum\limits_{k=1}^{K} \left[a_{k,n}\tilde{\tbf}_{k,n+1}+(1-a_{k,n})\tilde{\tbf}_{k,m}\right].
\label{eq:clu}
\end{align}
We summarize RERCE-Fed with continual local updates in Algorithm \ref{Alg:RFed_CLU}.

\begin{algorithm}[t!]
\textbf{Parameters}: penalty parameters $\rho$, number of clients $K$, set of clients $\Scal$ \\
\textbf{Initialization}: global model $ \wbf_0 =  \wbf_{-1} = \mathbf{0}$, local models $\wbf_{k,0} = \widehat{\wbf}_k$\\[1mm]
\textbf{For} \,$n=1,\cdots,$ \textit{Until Convergence} \\[1mm]
\hspace*{5mm} The server randomly selects a subset $\Scal_n$ of its clients and sends the aggregated global model $\sbf_n$ to them. \\[1mm]
\textbf{Client Local Update}:\\[1mm]
\hspace*{5mm} \textbf{If} $k \in \Scal_n$ \\[1mm]
\hspace*{10mm} Receive $\tilde{\sbf}_{k,n}$, a noisy version of $\sbf_{n}$, from the server. \\
\hspace*{10mm} Store the latest global model $\tilde{\sbf}_{k,m}=\tilde{\sbf}_{k,n}$.\\
\hspace*{10mm} Update the local model as\\[1mm] 
\hspace*{15mm} $\wbf_{k,n+1}=( \Ibf-\rho \Nbf_{k})\wbf_{k,n} +\rho \Nbf_{k} \tilde{\sbf}_{k,n}$ \\[1mm]
\hspace*{10mm} Send $\tbf_{k,n+1} = 2\wbf_{k,n+1}-\wbf_{k,n}$ to the server. \\[1mm]
\hspace*{5mm}\textbf{Else} \\[1mm]
\hspace*{10mm} Update the local model as\\ 
\hspace*{15mm} $\wbf_{k,n+1}=( \Ibf-\rho \Nbf_{k})\wbf_{k,n} +\rho \Nbf_{k} \tilde{\sbf}_{k,m}$ \\[1mm]
\hspace*{5mm}\textbf{EndIf} \\[1mm]
\textbf{Aggregation at the Server}:\\[1mm]
\hspace*{5mm} The server receives $\tilde{\tbf}_{k,n+1}$, noisy versions of the locally updated models from the selected clients $k \in \Scal_n$ and aggregates them with $\tilde{\tbf}_{k,m}$, the stored local model estimates of the non-selected clients via\\[1mm]
\hspace*{5mm} $\sbf_{n+1} =\frac{1}{K}\sum\limits_{k=1}^{K} \left[a_{k,n}\tilde{\tbf}_{k,n+1}+(1-a_{k,n})\tilde{\tbf}_{k,m}\right]$\\
\textbf{EndFor}\\
\caption{RERCE-Fed with continual local updates.} \label{Alg:RFed_CLU}
\end{algorithm} 


\section{Performance Analysis} \label{sec:analyis}

In this section, we analyze the performance of RERCE-Fed theoretically. We establish the convergence of the iterates $\wbf_{k,n}$ in both mean and mean-square senses as $n \rightarrow \infty$, even in the presence of noisy communication links. Given that $\wbf_n$ represents the average of client estimates $\wbf_{k,n}$, its convergence follows accordingly.

To facilitate the analysis, we define the extended optimal global model as $\wbf_e^{\star} = \mathbf{1}_{2K} \otimes \wbf^{\star}$ and the vector containing the client local model estimates as $$\wbf_{e,n} = \col \{ \wbf_{1,n}, \cdots, \wbf_{K,n}, \wbf_{1,n-1}, \cdots, \wbf_{K,n-1} \},$$ where $\mathbf{1}_{2K}$ is the $2K \times 1$ vector of all ones, $\otimes$ is the Kronecker product, and $\col \{\cdot\}$ denotes column-wise stacking.

Substituting \eqref{eq:FDxbar4} into \eqref{eq:FDx3}, the global recursion of the proposed algorithm can be stated as 
\begin{equation}
\wbf_{e,n+1} = \Acalbf_n \wbf_{e,n} + \boldsymbol{\zeta}_{n} + \boldsymbol{\eta}_{n},
\label{eq:conv}
\end{equation}
where
\begin{equation}
\Acalbf_n  = 
\left[
\begin{array}{c;{2pt/2pt}c}
    \Acalbf_{n,1} & \Acalbf_{n,2} \\ \hdashline[2pt/2pt]
    \Ibf & \mathbf{0}
\end{array}
\right],
\label{eq:conv1}
\end{equation}
and the extended noise vectors $\boldsymbol{\zeta}_{n}$ and $\boldsymbol{\eta}_{n}$ stack the vectors
\begin{equation}
    a_{k,n} \rho \Nbf_k \boldsymbol{\zeta}_{k,n}
\end{equation}
and
\begin{equation}
    a_{k,n} \frac{\rho}{\Ccal} \Nbf_k \sum\limits_{j=1}^{K} \left( 2 a_{j,n-1}  \boldsymbol{\eta}_{j,n-1} - a_{j,n-2}  \boldsymbol{\eta}_{j,n-2}  \right)
\end{equation}
at their top halves, respectively, and zeros at their bottom halves.
The value of $\Acalbf_n \in \Rbb^{2LK \times 2LK}$ depends on the iteration number $n$ as the server selects a random number of clients at each iteration. Its sub-matrices of size $LK \times LK$ are block matrices whose $L \times L$ blocks are calculated as
\begin{subequations} \label{eq:A_MAT_def}
\begin{align}
\left[ \Acalbf_{n,1} \right]_{ii} & =  \Ibf - a_{i,n} \rho \Nbf_i + 2 a_{i,n} a_{i,n-1} \frac{\rho}{\Ccal}\Nbf_i\\ 
\left[ \Acalbf_{n,1} \right]_{ij} & = 2 a_{i,n} a_{j,n-1} \frac{\rho}{\Ccal}\Nbf_i,\ i\neq j\\ 
\left[ \Acalbf_{n,2} \right]_{ij} & = - a_{i,n} a_{j,n-2} \frac{\rho}{\Ccal}\Nbf_i.
\label{eq:Amat}
\end{align}
\end{subequations}

To make the analysis tractable, we adopt the following assumptions:

\noindent A1: The communication noise vectors of both uplink and downlink, $\boldsymbol \eta_{k,n}$ and $\boldsymbol \zeta_{k, n}\ \forall k,n$, are independently and identically distributed. In addition, they are independent of each other and all other stochastic variables.

\noindent A2: The random scheduling variables, $a_{k,n}\ \forall k,n$, are independent and follow the same Bernoulli distribution with parameter $\Bar{a} = \frac{\Ccal}{K}$, i.e., $a_{k,n} = 1$ with probability $\Bar{a}$ and $a_{k,n} = 0$ with probability $1 - \Bar{a}$.

\subsection{Mean Convergence} \label{sec:mean_conv}

Taking the expected value of both sides of \eqref{eq:conv}, while considering A1-A2, yields
\begin{equation}
\Ebb [\wbf_{e,n+1}] = \bar{\Acalbf}\ \Ebb[\wbf_{e,n}], 
\label{eq:ea}
\end{equation}
where
\begin{equation}
\bar{\Acalbf} = \Ebb [\Acalbf_n] = 
\left[
\begin{array}{c;{2pt/2pt}c}
    \Ibf - \Ocalbf + 2 \Pcalbf & - \Pcalbf \\ \hdashline[2pt/2pt]
    \Ibf & \mathbf{0}
\end{array}
\right],
\label{eq:conv1_2}
\end{equation}
\begin{subequations} \label{eq:A_bar}
\begin{align} 
\Ocalbf & = \Bar{a} \rho\ \bdiag \{\Nbf_1,\cdots,\Nbf_K\}, \\ 
\Pcalbf & = \frac{\Bar{a} \rho}{K}\mathbf{1}^{\intercal}_{K} \otimes \bcol \{\Nbf_1,\cdots,\Nbf_K\},
\label{eq:Ncal}
\end{align}
\end{subequations}
and $\bcol\{\cdot\}$ and $\bdiag\{\cdot\}$ represent block column-wise stacking and block diagonalization, respectively.
Given that the initial estimate of each client is deterministic, \eqref{eq:ea} can be recursively unfolded over $n$ iterations as
\begin{equation}
\Ebb [\wbf_{e,n+1}]
= \bar{\Acalbf}^{n} \col \{\widehat{\wbf}_{e},\mathbf{0}\}, 
\label{eq19}
\end{equation}
where $\widehat{\wbf}_{e} = \col \{ \widehat{\wbf}_{1}, \cdots, \widehat{\wbf}_{K} \}$.

Considering \eqref{eq:conv1_2} and \eqref{eq:A_bar}, the matrix $\bar{\Acalbf}$ is right-stochastic, as its block rows add up to the identity matrix, and has a spectral radius of $1$ with the geometric and algebraic multiplicity of $L$ \cite[Lemma 6(a)]{4407653}.
Therefore, using the Jordan canonical form
$\bar{\Acalbf} = \Ubf\Jbf\Ubf^{-1}$, we have
%
\begin{equation}
\bar{\Acalbf}^{n} = \Ubf\Jbf^n\Ubf^{-1}
\end{equation}
and, as $n \rightarrow \infty$, we obtain
\begin{equation}
\bar{\Acalbf}^{\infty} = \sum_{i=1}^{L} \ubf_i \vbf_i^{\intercal}.
\end{equation}
where $\ubf_i$ and $ \vbf_i^{\intercal}$ denote the $i$th right and left eigenvectors of matrix $\bar{\Acalbf}$ corresponding to the eigenvalue $1$, respectively.

The $L$ dominant right eigenvectors (corresponding to the eigenvalue $1$) have the form 
$$ \ubf_i = \col\{\boldsymbol{\epsilon}_i, \cdots, \boldsymbol{\epsilon}_i\} \ \ \forall i \in \{ 1, \cdots, L \}, $$
where $\boldsymbol{\epsilon}_i$ denotes an $L$-dimensional vector with one in its $i$th entry and zero in all other entries~\cite[Lemma 6(b)]{4407653}.
Additionally, the $L$ dominant left eigenvectors (corresponding to the eigenvalue $1$) satisfy 
$$ \vbf_i^{\intercal} \bar{\Acalbf} = \vbf_i^{\intercal} \ \ \forall i \in \{ 1, \cdots, L \} $$ 
that results in
\begin{subequations} \label{eq:left_eig}
\begin{align}
    \vbf_{i,1}^{\intercal} (\Ibf - \Ocalbf + 2 \Pcalbf) + \vbf_{i,2}^{\intercal} & = \vbf_{i,1}^{\intercal} \\
    \vbf_{i,1}^{\intercal} \Pcalbf + \vbf_{i,2}^{\intercal} & = \mathbf{0}, \label{eq:left_eig_v2}
\end{align}
\end{subequations}
where $\vbf_i$ can be written as $\col \{ \vbf_{i,1}, \vbf_{i,2} \}$. 
Therefore, considering $\vbf_i^{\intercal} \ubf_i = 1$, we have
\begin{subequations} \label{eq:left_eig_cond}
\begin{align} 
    \vbf_{i,1}^{\intercal} (\Ibf - \Ocalbf + \Pcalbf) & = \vbf_{i,1}^{\intercal} \\
    \vbf_{i,1}^{\intercal} (\Ibf - \Pcalbf) \ubf_{i,1} & = 1,
\end{align}
\end{subequations}
where $\ubf_{i,1}$ contains the first $KL$ entries of $\ubf_i$. 

Following the same procedure as in \cite[Lemma 6(c)]{4407653}, $\vbf_{i,1}\ \forall i \in \{1,\cdots,L\}$ can be determined from \eqref{eq:left_eig_cond}. Subsequently, $\vbf_{i,2}$ can be computed using \eqref{eq:left_eig_v2}. Hence, we have 
\begin{subequations} \label{eq:left_ei_val}
\begin{align}
    \vbf_{i,1}^{\intercal} & = \frac{\boldsymbol{\epsilon}_i^{\intercal}}{2} \left( \sum_{k=1}^{K} \Xbf_k^{\intercal} \Wcalbf_k \Xbf_k \right)^{-1} \left[\Nbf_1^{-1}, \cdots, \Nbf_K^{-1} \right] \\
    \vbf_{i,2}^{\intercal} & = - \vbf_{i,1}^{\intercal} \Pcalbf \ \ \forall i \in \{ 1, \cdots, L \}.
\end{align}
\end{subequations}
%
%
Consequently, in view of \cite[Proposition 4]{4407653}, taking the limit on both sides of \eqref{eq19} leads to
\begin{align}
& \lim_{n \rightarrow \infty} \Ebb \left[ \wbf_{e,n} \right]
= \sum_{i=1}^{L} \ubf_i \vbf_{i,1}^{\intercal} \widehat{\wbf}_{e} \\
& = \sum_{i=1}^{L} \ubf_i \boldsymbol{\epsilon}_i^{\intercal} \left[ \left( \sum_{k=1}^{K} \Xbf_k^{\intercal} \Wcalbf_k \Xbf_k \right)^{-1} \sum_{k=1}^{K} \Xbf_k^{\intercal} \Wcalbf_k \ybf_k \right] = \wbf_{e}^{\star}. \notag
\end{align}
Therefore, the RERCE-Fed algorithm is unbiased and the proof for mean convergence is complete. 

\subsection{Mean-Square Convergence} \label{sec:mean_square_conv}

Let us define the deviation vector as $\tilde{\wbf}_{e,n} = {\wbf}_{e,n} -\wbf_{e}^{\star}$. Since $\Acalbf_n$ is block right-stochastic, i.e., its block rows add up to the identity matrix, we have $\Acalbf_n \wbf_{e}^{\star} = \wbf_{e}^{\star}$. Therefore, by defining the weighted norm-square of $\tilde{\wbf}_{e,n}$ as $\|\tilde{\wbf}_{e,n}\|_{\boldsymbol\Sigma}^2 = \tilde{\wbf}^{\intercal}_{e,n} {\boldsymbol \Sigma} \tilde{\wbf}_{e,n}$, where $\boldsymbol \Sigma$ is an arbitrary positive semi-definite matrix, we obtain the variance relation as 
\begin{align}\label{varrel}
\Ebb & \left[ \|\tilde{\wbf}_{e,n+1}\|_{\boldsymbol \Sigma}^2 \right] \notag \\
& = \Ebb \left[ \|\tilde{\wbf}_{e,n}\|_{\boldsymbol \Sigma^{\prime}}^2 \right] + \Ebb \left[ \boldsymbol{\zeta}^{\intercal}_{n} {\boldsymbol \Sigma} {\boldsymbol \zeta}_{n} \right] + \Ebb \left[ \boldsymbol \eta_{n}^{\intercal} {\boldsymbol \Sigma} \boldsymbol \eta_{n} \right],
\end{align}
where the cross terms vanish as they are independent and uncorrelated with one another and all other variables. The matrix ${\boldsymbol \Sigma^{\prime}}$ is given by
\begin{align}\label{sigmaprime}
{\boldsymbol \Sigma^{\prime}} = \Ebb \left[ {\Acalbf}_{n}^{\intercal} {\boldsymbol \Sigma} {\Acalbf}_{n} \right].
\end{align}

We can describe the relationship between $\boldsymbol{\Sigma}^{\prime}$ and $\boldsymbol{\Sigma}$ as $\boldsymbol{\sigma^{\prime}} = \bvec\{\boldsymbol{\Sigma}^{\prime}\} = {\Qcalbf}\boldsymbol{\sigma}$, where ${\Qcalbf} = \Ebb[{\Acalbf}_{n}^\intercal\otimes_b{\Acalbf}_{n}^\intercal]$ and $\boldsymbol{\sigma}=\bvec\{\boldsymbol{\Sigma}\}$. Here, $\otimes_b$ is the block Kronecker product (Tracy–Singh product \cite{tracy1972new}) and $\bvec\{\cdot\}$ denotes the block vectorization operation \cite{koning1991block}. We evaluate ${\Qcalbf}$ in Appendix~\ref{EvalQA}.

Using the relationship between the matrix trace operator, denoted by $\tr(\cdot)$, and the block Kronecker product, we evaluate the second term on the right-hand side (RHS) of \eqref{varrel} as
\begin{align}
\Ebb[\boldsymbol{\zeta}^{\intercal}_{n} {\boldsymbol \Sigma} {\boldsymbol \zeta}_{n}]
= \tr \big(\Ebb[\boldsymbol{\zeta}_{n}{\boldsymbol \zeta}^{\intercal}_{n}]\boldsymbol \Sigma \big)
= \boldsymbol \phi^{\intercal} \boldsymbol{\sigma},
\end{align}
where 
$\boldsymbol\phi = \bvec\{\Ebb[\boldsymbol{\zeta}_{n}{\boldsymbol \zeta}^{\intercal}_{n}]\}$, 
\begin{equation} \label{eq:E_zeta}
\Ebb [ \boldsymbol{\zeta}_{n} {\boldsymbol \zeta}^{\intercal}_{n} ]
= \bar{a} \rho^2 \bdiag \{\sigma^2_{\zeta_1} \Nbf_{1}^{2},\cdots,\sigma^2_{\zeta_K} \Nbf_{K}^{2}, \underbrace{\mathbf{0}, \cdots, \mathbf{0}}_{K} \},
\end{equation}
and $\sigma^2_{\zeta_k}$ is the variance of the downlink noise for client $k$. 

Following a similar procedure, we evaluate the third term on the RHS of \eqref{varrel} as
\begin{align}
\Ebb[ \boldsymbol \eta_{n}^{\intercal} {\boldsymbol \Sigma} \boldsymbol \eta_{n}]
= \tr \big(\Ebb[\boldsymbol{\eta}_{n}{\boldsymbol \eta}^{\intercal}_{n}]\boldsymbol \Sigma \big) 
= \boldsymbol \varphi^{\intercal} \boldsymbol{\sigma},
\end{align}
where $\boldsymbol \varphi = \bvec \{ \Ebb[\boldsymbol{\eta}_{n}{\boldsymbol \eta}^{\intercal}_{n}] \}$,
\begin{equation} \label{eq:E_eta}
    \Ebb [ \boldsymbol{\eta}_{n} {\boldsymbol \eta}^{\intercal}_{n} ]
    = \frac{5 \rho^2}{K^2} \sum_{k=1}^{K} \sigma_{\eta_k}^2 \bdiag \{ \Nbf_{1}^{2},\cdots, \Nbf_{K}^{2}, \underbrace{\mathbf{0}, \cdots, \mathbf{0}}_{K}\},
\end{equation}
and $\sigma_{\eta_k}^2$ is the variance of the uplink noise for client $k$. Note that in \eqref{eq:E_eta}, $\Ebb [ a_{i,n} a_{j,n} ] \ll 1 \ \forall i \neq j$, and can be neglected.

Utilizing the above results, we can write the global recursion for the weighted mean-squared error (MSE) of RERCE-Fed as
\begin{align}\label{globrec}
\Ebb &\left[ \|\tilde{\wbf}_{e,n+1} \|_{\bvec^{-1} \{ \boldsymbol \sigma \}}^2 \right] \notag \\ 
& = \Ebb \left[ \|\tilde{\wbf}_{e,n}\|_{\bvec^{-1}\{{\Qcalbf}\boldsymbol \sigma \}}^2 \right] + \boldsymbol \phi^{\intercal} \boldsymbol{\sigma} + \boldsymbol \varphi^{\intercal} \boldsymbol{\sigma}.
\end{align}
By defining $\boldsymbol\psi = \boldsymbol \phi + \boldsymbol \varphi$ and iterating \eqref{globrec} backward to $n=1$, we obtain
\begin{align}\label{varrel2}
\Ebb &\left[ \|\tilde{\wbf}_{e,n+1}\|_{\bvec^{-1}\{ \boldsymbol \sigma \}}^2 \right] \notag \\
& = \Ebb \left[ \|\tilde{\wbf}_{e,1}\|_{\bvec^{-1}\{{\Qcalbf}^{n} \boldsymbol \sigma \}}^2 \right] + \boldsymbol \psi^{\intercal} \sum_{i=0}^{n-1} {\Qcalbf}^i \boldsymbol \sigma.
\end{align}  
Using the Jordan canonical form of ${\Qcalbf}$, we have
\begin{align}
    {\Qcalbf}^i = \Ucalbf \Jcalbf^i \Ucalbf^{-1} = \sum_{\ell = 1}^{4L^2K^2} \mu_{\ell}^{i} \boldsymbol{u}_{\ell} \boldsymbol{v}_{\ell}^{\intercal},
\end{align}
where $\mu_{\ell}$, $\boldsymbol{u}_{\ell}$ and $\boldsymbol{v}_{\ell}$ denote the $\ell$th eigenvalue of $\Qcalbf$ and its corresponding right and left eigenvectors, respectively. In Appendix~\ref{spectQA}, we show that the spectral radius of $\Qcalbf$ is one with the geometric and algebraic multiplicity of $L^2$, i.e., $\mu_{\ell} = 1 \ \forall \ell \in \{1,\dots, L^2\}$.



\textbf{Proposition 1:} $\forall \ell \in \{1,\dots, L^2\}$, we have 
\begin{align}
    \boldsymbol \psi^{\intercal} \boldsymbol{u}_{\ell} \boldsymbol{v}_{\ell}^{\intercal} \boldsymbol \sigma = 0.
\end{align}

\textit{Proof:} See Appendix \ref{proof_prop1}.

\subsection{Steady-State Mean-Square Error} \label{sec:mean_square_error}

Setting $\boldsymbol \sigma = \bvec \{ \Ibf_{2KL} \}$, letting $n \rightarrow \infty$ on both sides of \eqref{varrel2}, and using Proposition 1, we compute the steady-state MSE of RERCE-Fed, denoted by $\Ecal$, as
\begin{align} \label{SSMSE}
    \Ecal & = \lim_{n \to \infty} \Ebb \left[ \tilde{\wbf}_{e,n}^{\intercal} \tilde{\wbf}_{e,n} \right] \notag \\
    & = \underbrace{ \tilde{\wbf}_{e,1}^{\intercal} \boldsymbol{\Sigma}_{\infty} \tilde{\wbf}_{e,1} }_{\Ecal_\nu} + \underbrace{ \sum_{\ell = L^2 + 1}^{4L^2K^2} \left( 1 - \mu_{\ell} \right)^{-1} \boldsymbol \psi^{\intercal} \boldsymbol{u}_{\ell} \boldsymbol{v}_{\ell}^{\intercal} \boldsymbol \sigma}_{\Ecal_\psi}, 
\end{align}
where
\begin{align} \label{eq:E_infity}
 \boldsymbol{\Sigma}_{\infty} & = \bvec^{-1} \left\{{\Qcalbf}^{\infty} \boldsymbol \sigma \right\}  \notag \\
 & = \bvec^{-1} \left\{ \sum \limits_{\ell = 1}^{L^2} \boldsymbol{u}_{\ell} \boldsymbol{v}_{\ell}^{\intercal} \bvec \{ \Ibf_{2KL} \} \right\}.
\end{align}

\textit{Remark 1:} The observation noise $\boldsymbol\nu_k$ in \eqref{eq:ModelFL} and the initial client estimates $\wbf_{k,0}$ affect the first term on the RHS of~\eqref{SSMSE}, $\Ecal_\nu$, resulting in a noise floor. Additionally, scheduling parameters, including the client participation probability and the number of participating clients in each iteration, along with the penalty parameter, the number of clients, the local data at each client, and the variance of noise in the uplink and downlink communications impact the second term on the RHS of~\eqref{SSMSE}, $\Ecal_\psi$. We will explore these factors further through simulations in section \ref{sec:ill_square}.

\section{Simulation Results} \label{sec:simulation}

In this section, we conduct several numerical experiments to evaluate the performance of the proposed algorithm and validate our theoretical findings. We consider a federated network consisting of $K$ clients. Each client has non-IID data $\{ \Xbf_k,\ybf_k\}$, where the entries of $\Xbf_k$ are drawn from a normal distribution $\Ncal(\mu_{k}, \sigma_{k}^2)$ with $\mu_{k} \in \Ucal(-0.5,0.5)$ and $\sigma_{k}^2 \in \Ucal(0.5,1.5)$. Here, $\Ucal(\cdot,\cdot)$ denotes a uniform distribution with the specified lower and upper bounds. The local data size for each client is randomly selected from a uniform distribution, i.e., $d_k \in \Ucal(50,90)$. We set the weight matrices at each client $k$ to the inverse covariance matrix of $\ybf_k$, i.e., $ \Wcalbf_k= \Ebb [ \left( \ybf_k - \Ebb [\ybf_k] \right) \left( \ybf_k - \Ebb [\ybf_k] \right)^{\intercal}]^{-1} $. 

The local data is related as per~\eqref{eq:ModelFL} given the model parameter vector $\boldsymbol{\omega}$ with its entries drawn from a normal distribution $\Ncal(0,1)$. The observation noise $\boldsymbol{\nu}_k$ for each client is zero-mean IID Gaussian with variance $\sigma_{\nu_k}^2$. The additive noise in both the uplink and downlink channels is zero-mean IID Gaussian with variances $\sigma^2_{\eta_k}$ and $\sigma^2_{\zeta_k}$, respectively. We set the penalty parameter as $\rho = 1$. The server randomly selects a subset of $\Ccal$ clients with uniform probability in each iteration.
We evaluate the algorithm performance on the client side via the normalized MSE (NMSE) defined at iteration $n$ as 
\begin{equation} 
    \frac{1}{K} \sum\limits_{k=1}^{K} \frac{  \| \wbf_{k,n}  - \wbf^{\star} \|_2^2 }{\| \wbf^{\star} \|_2^2}.
\label{eq:MSDST}
\end{equation}
We average the NMSE learning curves over $100$ independent trials to obtain our simulation results.

We present the results of our numerical experiments in four subsections. In the first subsection, we evaluate the performance of the proposed RERCE-Fed algorithm in comparison to its predecessors. In the second subsection, we evaluate the performance of RERCE-Fed with continual local updates and compare it with RERCE-Fed without continual local updates. 
In the last subsection, we corroborate our theoretical findings by demonstrating mean convergence and comparing theoretical predictions with simulation results, confirming the accuracy of our theoretical expression for MSE of RERCE-Fed.

\subsection{Performance of RERCE-Fed}

\begin{figure}[t!]
\centering
\includegraphics[width=.475\textwidth]{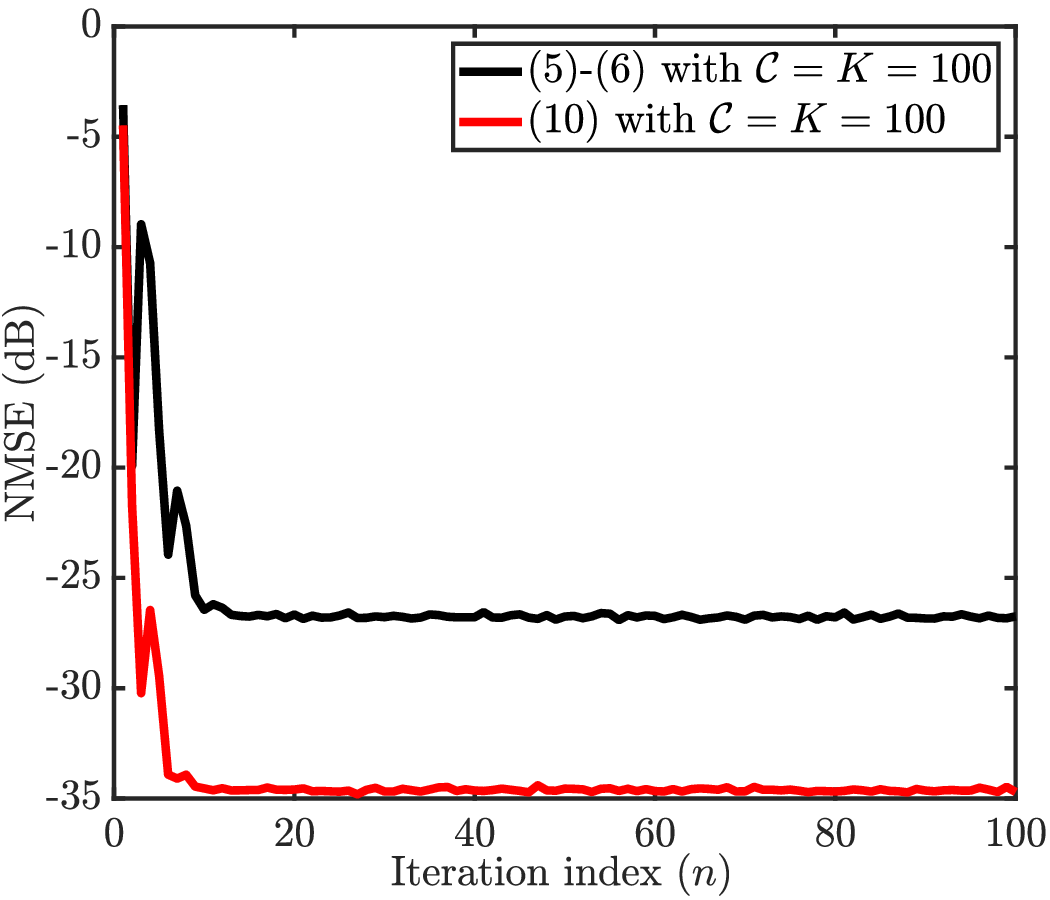}
\caption{NMSE of~\eqref{eq:FD}-\eqref{eq:FDxbar1} and~\eqref{fl3} for $\Ccal = K = 100$.}
\label{fig:SSP1}
\end{figure}

In our first experiment, we simulate the algorithms described by \eqref{eq:FD}-\eqref{eq:FDxbar1} and \eqref{fl3} to solve the considered WLS problem. We run these simulations with $K = 100$ clients, model parameter vector size of $L = 128$, and link noise variance of $\sigma^2_{\eta_k} = \sigma^2_{\zeta_k} = 6.25 \times 10^{-4}$. All clients participate in the FL process, i.e., $\Ccal = K = 100$. The corresponding learning curves are shown in Fig.~\ref{fig:SSP1}. We observe that \eqref{fl3} exhibits a $7\dB$ improvement over \eqref{eq:FD}-\eqref{eq:FDxbar1} in the presence of noisy communication links when all clients are involved in each iteration of the FL process.

\begin{figure}[t!]
\centering
\includegraphics[width=.475\textwidth]{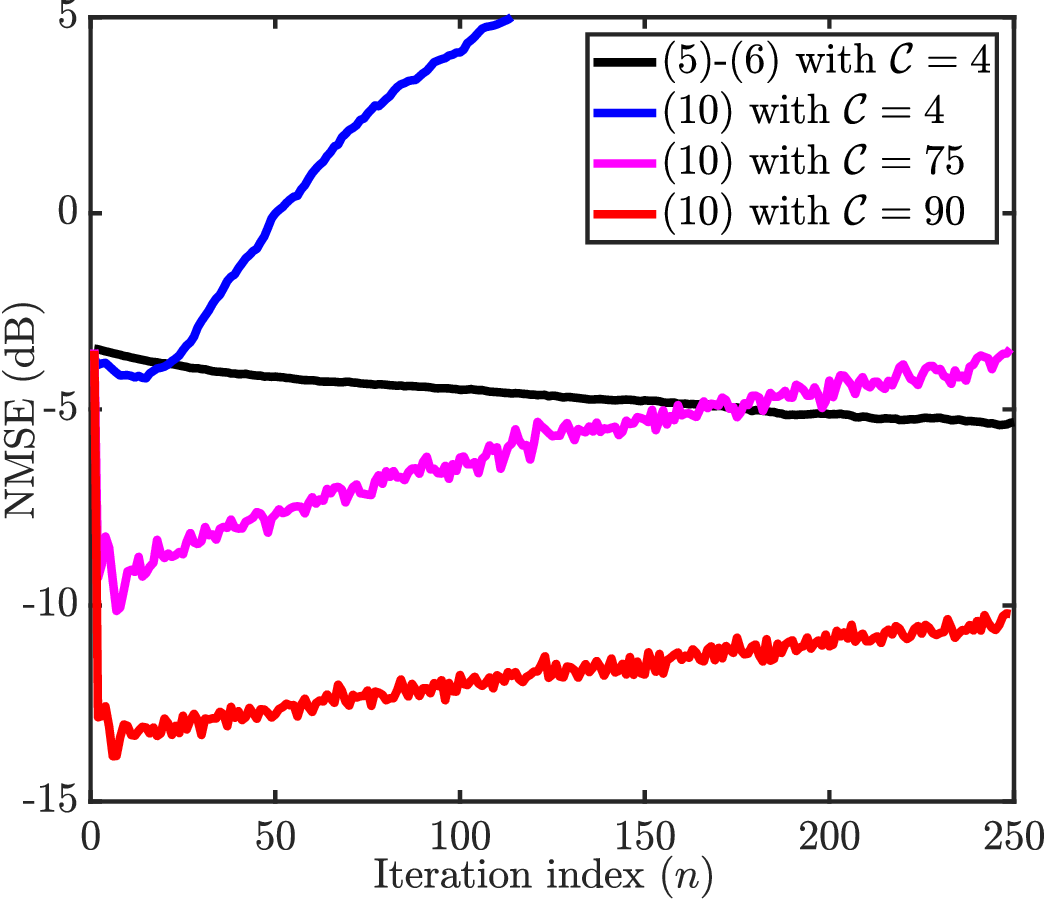}
\caption{NMSE of~\eqref{eq:FD}-\eqref{eq:FDxbar1} with $\Ccal = 4$ and \eqref{fl3} with $\Ccal \in \{4,75,90\}$.}
\label{fig:SSP2}
\end{figure}

In our second experiment, we examine the performance of \eqref{eq:FD}-\eqref{eq:FDxbar1} and \eqref{fl3} under similar conditions as the first experiment, but with the server randomly selecting a subset of clients to participate in each iteration. We simulate \eqref{eq:FD}-\eqref{eq:FDxbar1} with $\Ccal = 4$ and \eqref{fl3} with $\Ccal \in \{4,75,90\}$. The corresponding learning curves are shown in Fig.~\ref{fig:SSP2}. Unlike the first experiment, Fig.~\ref{fig:SSP2} illustrates that \eqref{fl3} fails to converge due to error accumulation, even when a majority of clients participate in every FL round, as observed for $\Ccal \in \{75,90\}$. Additionally, the performance of \eqref{eq:FD}-\eqref{eq:FDxbar1} degrades significantly when only a small subset of clients participate in each FL round. Consequently, both \eqref{eq:FD}-\eqref{eq:FDxbar1} and \eqref{fl3} exhibit an inability to cope with additive noise in communication links when the server selects only a subset of clients in each iteration.


\begin{figure}[t!]
\centering
\includegraphics[width=.475\textwidth]{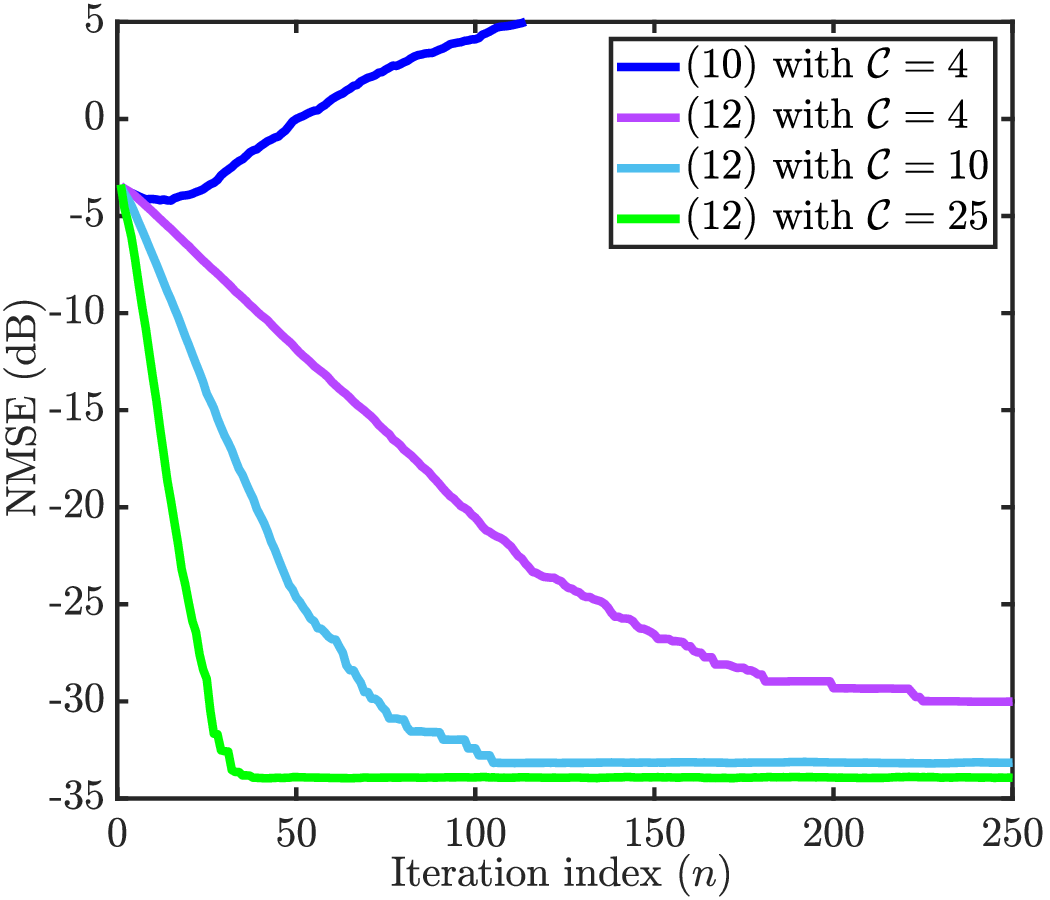}
\caption{NMSE of \eqref{fl3} with $\Ccal = 4$ and RERCE-Fed \eqref{eq:rercefed} for different numbers of participating clients $\Ccal \in \{4,10,25\}$.}
\label{fig:SSP3}
\end{figure}

In our third experiment, we assess the performance of the proposed RERCE-Fed algorithm in the presence of link noise and random client scheduling by the server to enhance communication efficiency. We simulate RERCE-Fed with $K = 100$ clients, $L = 128$, link noise variance of $\sigma^2_{\eta_k} = \sigma^2_{\zeta_k} = 6.25 \times 10^{-4}$, and different numbers of participating clients $\Ccal \in \{4,10,25\}$. The corresponding learning curves are depicted in Fig.~\ref{fig:SSP3}. As shown in Fig.~\ref{fig:SSP3}, RERCE-Fed exhibits robustness against communication noise, even when only a subset of clients participate in each FL round, a contrast to the results observed with \eqref{eq:FD}-\eqref{eq:FDxbar1} and \eqref{fl3} in the second experiment. Another important observation from Fig.~\ref{fig:SSP3} is the trade-off between the number of participating clients $\Ccal$ and the performance and convergence rate of RERCE-Fed. Specifically, increasing the number of participating clients leads to faster convergence and lower NMSE. However, this benefit diminishes once $\Ccal \geq 10$, as the performance of RERCE-Fed with $\Ccal \geq 10$ approaches that of the scenario where all clients participate in each FL round. This implies that RERCE-Fed can achieve accurate model parameter estimation while making more efficient use of available communication resources, even in the presence of noisy communication links.

\subsection{Performance of RERCE-Fed with Continual Local Updates}

\begin{figure}[t!]
\centering
\includegraphics[width=.475\textwidth]{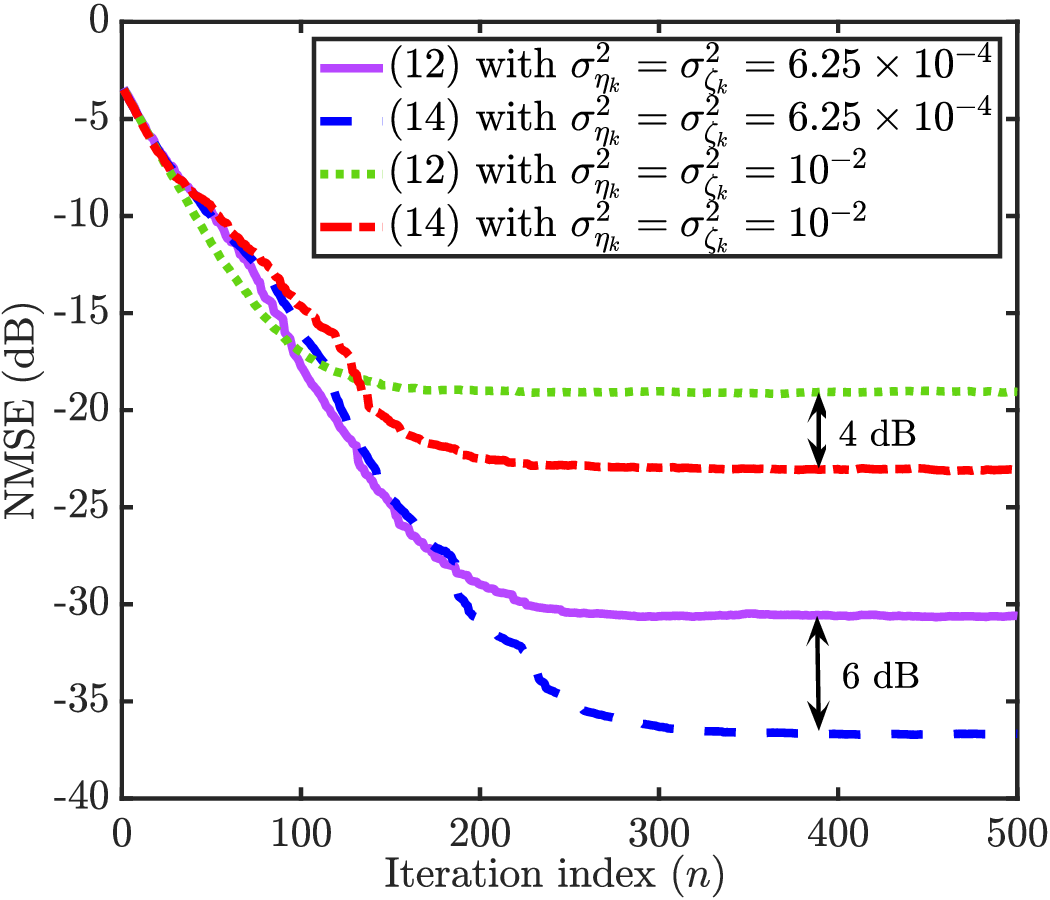}
\caption{NMSE of RERCE-Fed \eqref{eq:rercefed} and RERCE-Fed with continual local updates \eqref{eq:clu} for $\Ccal = 4$ and different uplink and downlink noise variances $ \sigma^2_{\eta_k} = \sigma^2_{\zeta_k} \in \{ 6.25 \times 10^{-4},10^{-2} \} $.}
\label{fig:APSIPA1}
\end{figure}
\begin{figure}[t!]
\centering
\includegraphics[width=.475\textwidth]{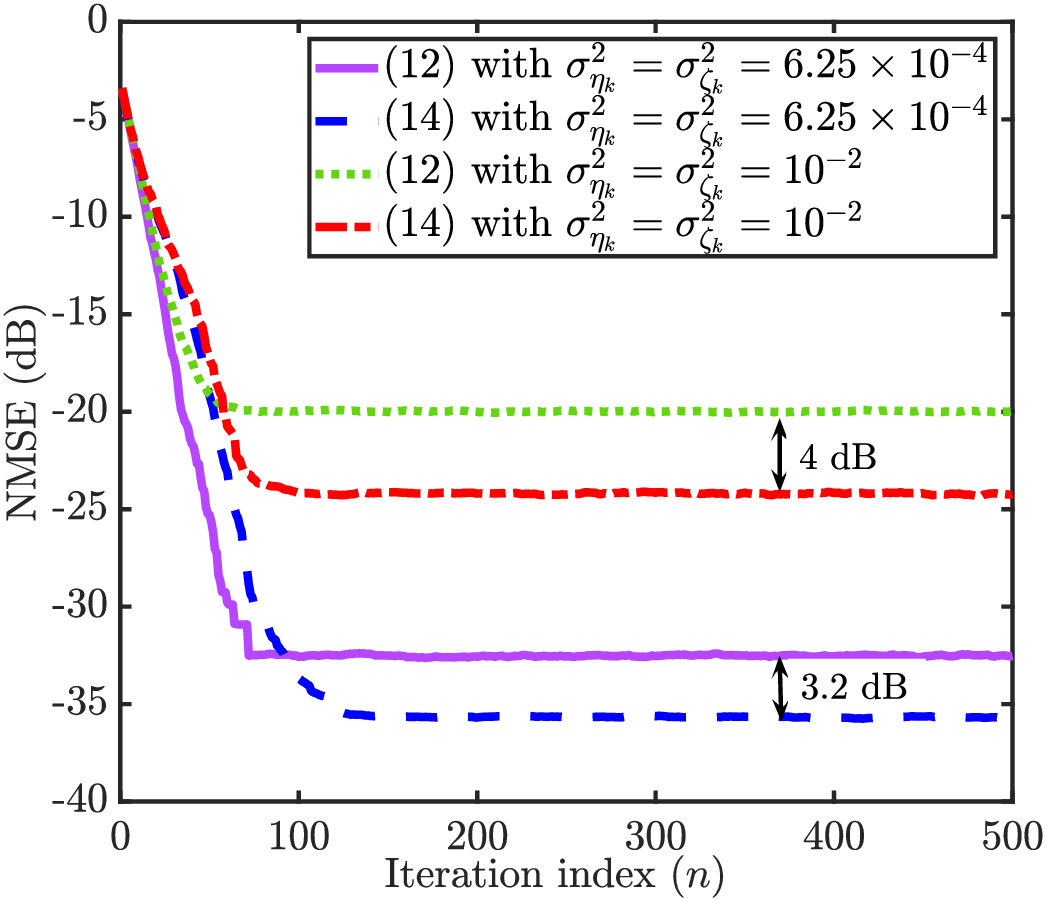}
\caption{NMSE of RERCE-Fed \eqref{eq:rercefed} and RERCE-Fed with continual local updates \eqref{eq:clu} for $\Ccal = 10 $ and different uplink and downlink noise variances $ \sigma^2_{\eta_k} = \sigma^2_{\zeta_k} \in \{ 6.25 \times 10^{-4},10^{-2} \} $.}
\label{fig:APSIPA2}
\end{figure}
\begin{figure}[t!]
\centering
\includegraphics[width=.475\textwidth]{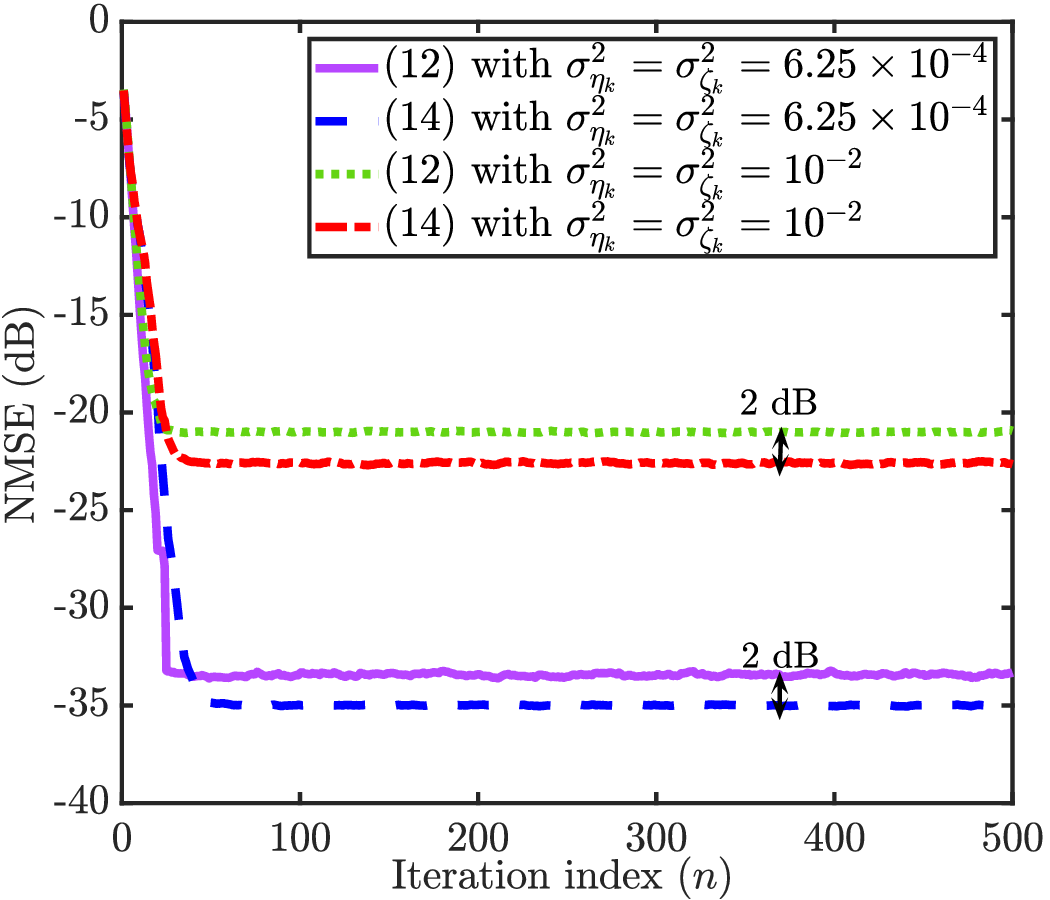}
\caption{NMSE of RERCE-Fed \eqref{eq:rercefed} and RERCE-Fed with continual local updates \eqref{eq:clu} for $\Ccal = 25$ and different uplink and downlink noise variances $ \sigma^2_{\eta_k} = \sigma^2_{\zeta_k} \in \{ 6.25 \times 10^{-4},10^{-2} \} $.}
\label{fig:APSIPA3}
\end{figure}

In our fourth experiment, we compare the performance of RERCE-Fed with and without continual local updates by plotting their corresponding learning curves given different numbers of participating clients and link noise variances. We present the results in Figs.~\ref{fig:APSIPA1}-\ref{fig:APSIPA3}. The number of clients selected at each iteration is $\Ccal \in \{4,10,25\}$, and the link noise variances are $ \sigma^2_{\eta_k} = \sigma^2_{\zeta_k} \in \{ 6.25 \times 10^{-4},10^{-2} \}$. From these figures, we observe that RERCE-Fed with continual local updates consistently exhibits robustness against communication noise, even when only a small subset of clients participate in every FL round. In addition, RERCE-Fed with continual local updates significantly outperforms its counterpart without continual local updates in all considered scenarios. These results indicate that allowing clients to continually update their local models, even when not selected by the server, leads to a substantial improvement in steady-state NMSE without adversely impacting the convergence rate. Additionally, as anticipated, increasing the link noise variance results in performance degradation.

\subsection{Comparison of Theory and Experiment} \label{sec:ill_square}

\begin{figure}[t!]
\centering
\includegraphics[width=.475\textwidth]{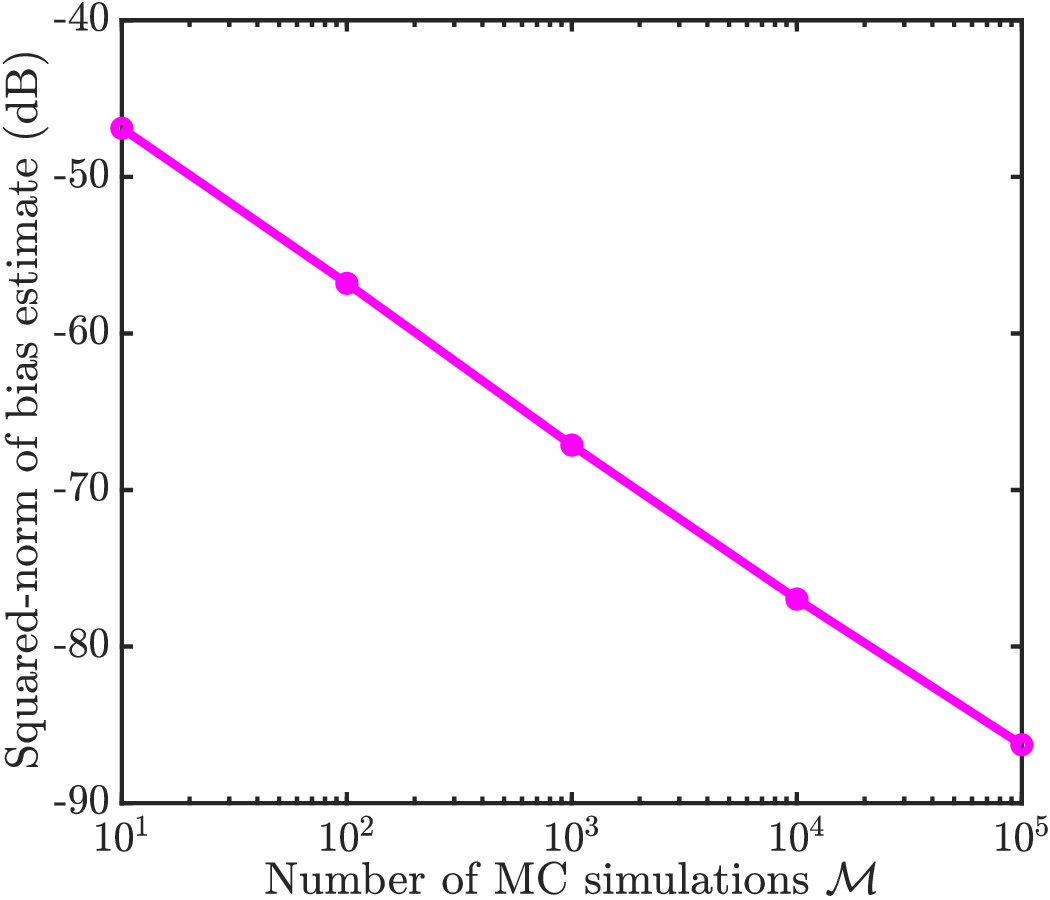}
\caption{Squared-norm of bias estimate of RERCE-Fed \eqref{eq:rercefed} with $K = 6$, $L = 6$, $\Ccal = 3$, and $\sigma^2_{\eta_k} = \sigma^2_{\zeta_k} = 10^{-4}$ for different numbers of MC runs $\Mcal \in \{ 10, 10^2, 10^3 , 10^4, 10^5 \}$.}
\label{fig:SE3}
\end{figure}
\begin{figure}[t!]
\centering
\includegraphics[width=.475\textwidth]{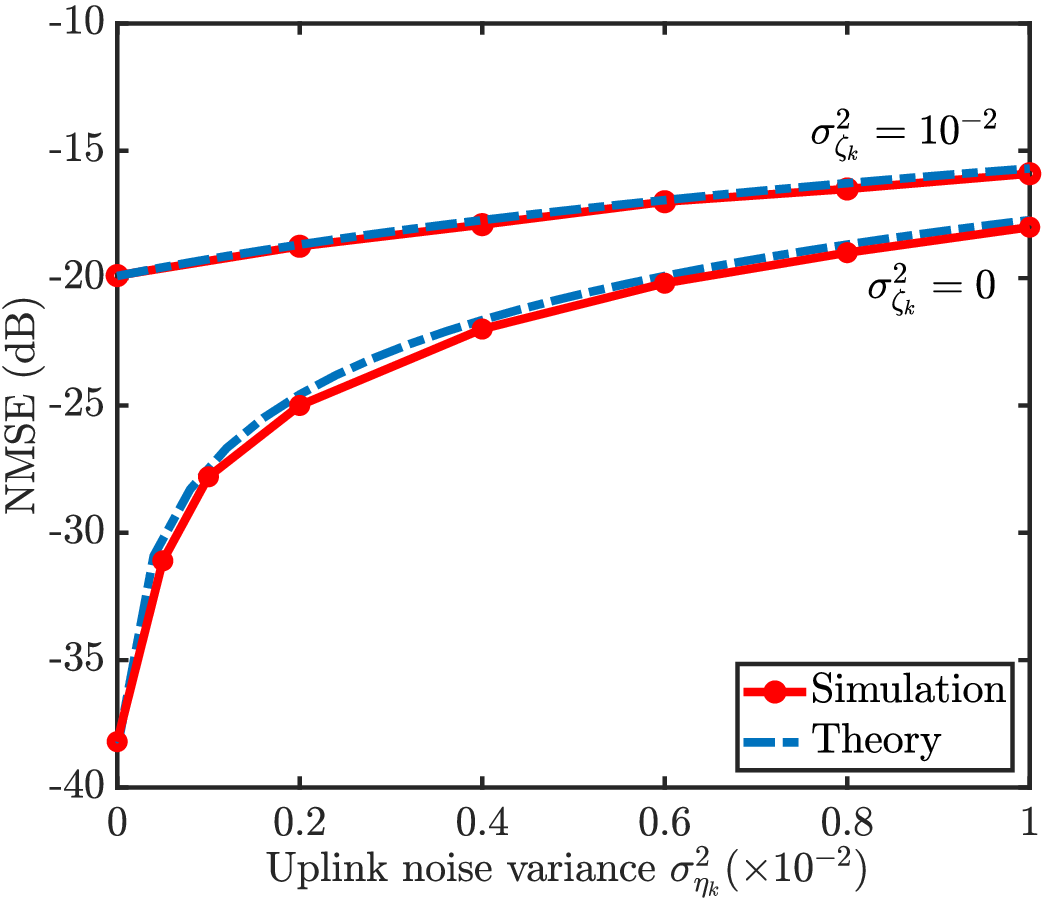}
\caption{NMSE of RERCE-Fed \eqref{eq:rercefed} with $\Ccal = 3$ for different uplink noise variances.}
\label{fig:DownNoise1}
\end{figure}
\begin{figure}[t!]
\centering
\includegraphics[width=.475\textwidth]{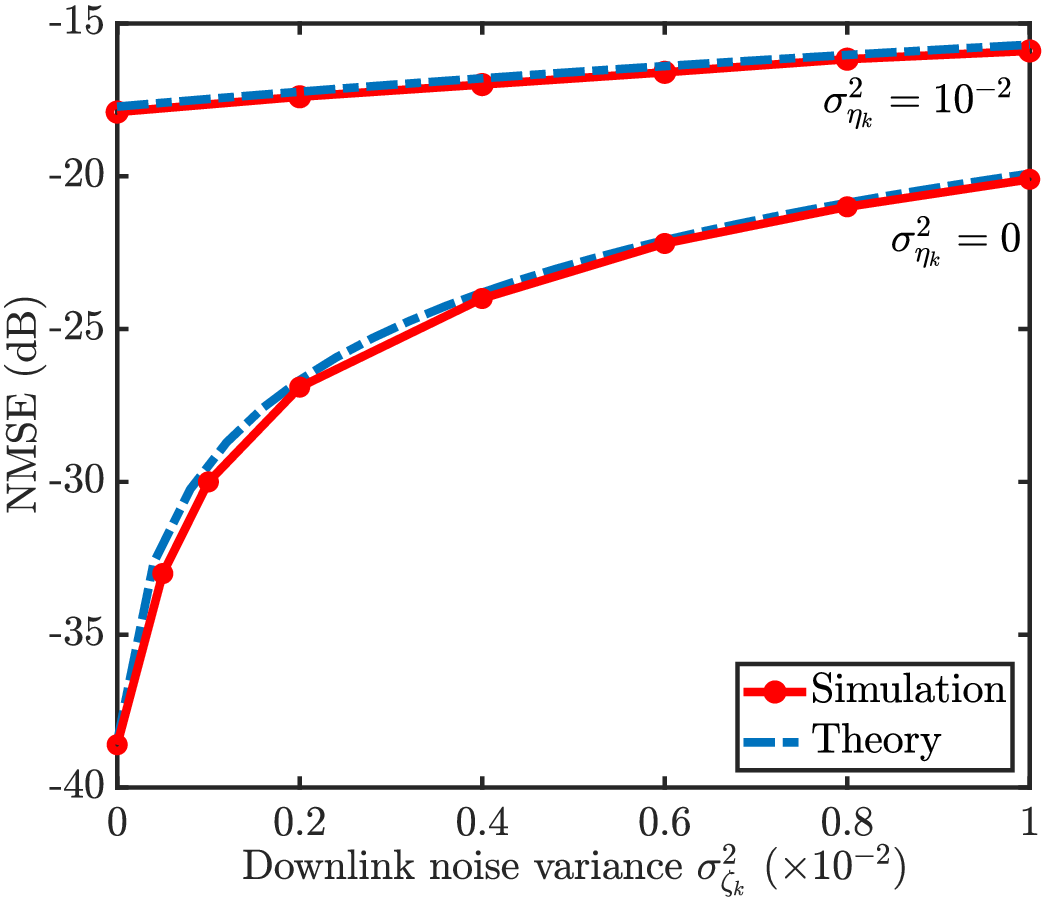}
\caption{NMSE of RERCE-Fed \eqref{eq:rercefed} with $\Ccal = 3$ for different downlink noise variances.}
\label{fig:UpNoise1}
\end{figure}
\begin{figure}[t!]
\centering
\includegraphics[width=.475\textwidth]{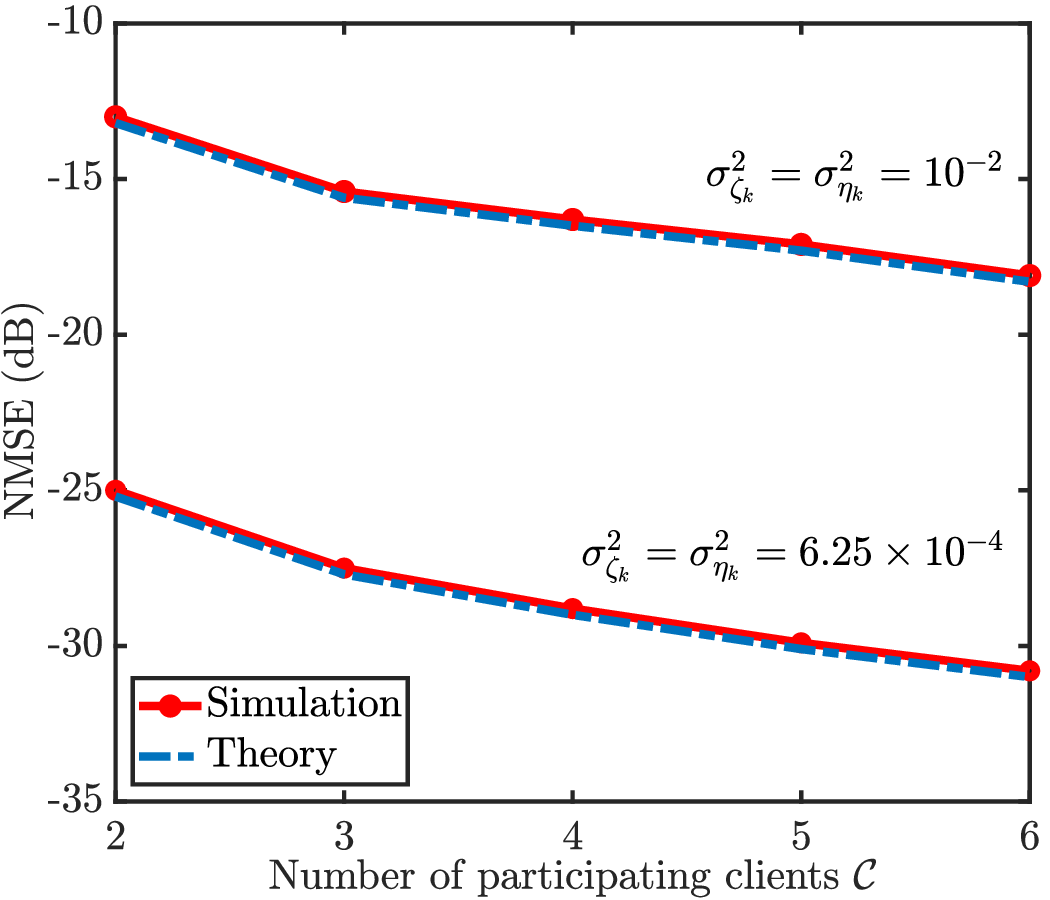}
\caption{NMSE of RERCE-Fed \eqref{eq:rercefed} for different numbers of participating clients $\Ccal \in \{ 2,3,4,5,6 \}$ and different link noise variances.}
\label{fig:DiffClient}
\end{figure}

In our fifth experiment, we demonstrate the mean convergence of RERCE-Fed. We simulate RERCE-Fed with $K = 6$ clients, $L = 6$, and link noise variances $\sigma^2_{\eta_k} = \sigma^2_{\zeta_k} = 10^{-4}$. In addition, the server randomly selects $\Ccal = 3$ clients in each iteration. In Fig.~\ref{fig:SE3}, we plot the squared $\ell_2$-norm of the global model parameter bias estimated for different numbers of Monte Carlo (MC) runs, denoted by $\Mcal$, specifically $\frac{1}{L} \| \frac{1}{\Mcal} \sum_{i=1}^{\Mcal} \wbf_n^{(i)} - \wbf^{\star} \|_2^2$, where $\wbf_n^{(i)}$ is the global model parameter estimate of the $i$th MC run. We can observe from Fig.~\ref{fig:SE3} that the squared-norm of the multivariate bias estimate decreases log-linearly with the number of MC runs, indicating the unbiasedness shown in section \ref{sec:mean_conv}. 

In our sixth experiment, we validate the accuracy of our theoretical expression for the steady-state MSE of RERCE-Fed in \eqref{eq:E_infity} and explore the impact of varying uplink and downlink noise variances, $\sigma^2_{\eta_k}$ and $\sigma^2_{\zeta_k}$, on performance. We simulate RERCE-Fed with $K = 6$ clients, $L = 6$ parameters, and different values for uplink and downlink noise variances. The server randomly selects $\Ccal = 3$ clients in each iteration. We present the theoretical predictions of steady-sate NMSE using \eqref{SSMSE} alongside the corresponding experimental values in Figs.~\ref{fig:DownNoise1}-\ref{fig:UpNoise1} as functions of $\sigma^2_{\eta_k}$, $\sigma^2_{\zeta_k}$, and $\Ccal$. The results show a close alignment between theory and experiment.
Furthermore, we observe an upward trend in the steady-state NMSE as either uplink or downlink noise variance increases. 

The uplink and downlink noise variances exhibit distinct effects depending on the number of participating clients $\Ccal$. 
While the error induced by uplink noise remains constant, the impact of downlink noise intensifies with an increasing number of participating clients. This observation is consistent with the intuition that averaging the model parameter estimates at the server can mitigate the adverse effect of uplink noise on the performance of RERCE-Fed. However, changing $\Ccal$ also affects $\Qcalbf$. To obtain a better understanding, we investigate the impact of $\Ccal$ on the performance of RERCE-Fed in our final experiment, where we consider $K = 6$, $L = 6$, link noise variances $\sigma^2_{\eta_k} = \sigma^2_{\zeta_k} \in \{6.25 \times 10^{-4}, 10^{-2}\}$, and $\Ccal \in \{ 2,3,4,5,6 \}$. The results presented in Fig.~\ref{fig:DiffClient} illustrate that increasing the number of participating clients improves the performance of RERCE-Fed. However, this improvement comes at the cost of higher resource utilization on the client side, which can be limiting in real-world FL scenarios.

\section{Conclusion and Future Work} \label{sec:conc}

We proposed RERCE-Fed, an FL algorithm designed to effectively reduce communication load while maintaining robustness against additive communication noise or errors. By employing ADMM to solve the WLS problem, we introduced a new local model update at the clients. This innovative solution mitigates the effects of communication noise without imposing additional computational burden on clients. Furthermore, we improved the communication efficiency by randomly selecting a subset of clients to participate in each learning round. To further optimize performance, we enabled non-selected clients to continue with their local updates, resulting in a modified version of RERCE-Fed. Our theoretical analysis confirmed the convergence of RERCE-Fed in both mean and mean-square sense, even with random client scheduling and communication over noisy communication links. In addition, we derived a closed-form expression for the steady-state MSE of the RERCE-Fed algorithm. Comprehensive numerical analysis substantiated our theoretical findings and confirmed the accuracy of our theoretical predictions. 
As a future extension of this work, we will investigate the impact of various adversarial attacks on the performance of RERCE-Fed, including model poisoning, data poisoning, and label poisoning attacks.

\appendices 

\section{Evaluation of $\mathbfcal{Q}$} \label{EvalQA}

Let us define
\begin{align*} 
\Acalbf_n = 
\begin{bmatrix} 
\Abf_{1, 1, n} &~ \Abf_{1, 2, n} &~ {\dots}&~ \Abf_{1, 2K, n} \\ 
\Abf_{2, 1, n} &~ \Abf_{2, 2, n} &~ {\dots}&~ \Abf_{2, 2K, n} \\ 
\vdots &~ \vdots &~ \ddots &~ \vdots \\ 
\Abf_{2K, 1, n} &~ \Abf_{2K, 2, n} &~ {\dots}&~ \Abf_{2K, 2K, n} 
\end{bmatrix},
\end{align*}
where
\begin{align*} 
\Abf_{i, j, n} = 
\begin{cases}
\Ibf_L - a_{i,n} \rho \Nbf_i                                 & i = j \in \{1, \ldots, K\} \\
\hspace{2mm} + 2 a_{i,n} a_{i,n-1} \frac{\rho}{\Ccal} \Nbf_i & \\
2 a_{i,n} a_{j,n-1} \frac{\rho}{\Ccal} \Nbf_i                & i \neq j \in \{1, \ldots, K\} \\
- a_{i,n} a_{j-K,n-2} \frac{\rho}{\Ccal} \Nbf_i              & i \in \{1, \ldots, K\}, \\
                                                             &  j \in \{K+1, \ldots, 2K\} \\
\Ibf_L                                                       & i \in \{K+1, \ldots, 2K\}, \\
                                                             & j \in \{1, \ldots, K\} \\
\mathbf{0}                                                   & {\text{otherwise}}.
\end{cases}
\end{align*}
Thus, $\Qcalbf^\intercal$ is given by
\begin{align*}
 \Ebb & \left [ {\Acalbf_n \otimes_{b} \Acalbf_n } \right] \\
& = \Ebb 
\begin{bmatrix} 
\Abf_{1, 1, n} \otimes_{b} \Acalbf_n & {\dots}& \Abf_{1, 2K, n} \otimes_{b} \Acalbf_n \\
\Abf_{2, 1, n} \otimes_{b} \Acalbf_n & {\dots}& \Abf_{2, 2K, n} \otimes_{b} \Acalbf_n \\
\vdots & \ddots & \vdots \\
\Abf_{2K, 1, n}\otimes_{b} \Acalbf_n & {\dots}& \Abf_{2K, 2K, n} \otimes_{b} \Acalbf_n 
\end{bmatrix},
\end{align*}
where
\begin{align*}
& \Ebb \left[ \Abf_{i, j, n} \otimes _{b} \Acalbf_n \right]\\
& = \Ebb 
\begin{bmatrix} 
\Abf_{i, j, n} \otimes \Abf_{1, 1, n} & {\dots} & \Abf_{i, j, n} \otimes \Abf_{1, 2K, n} \\ 
\Abf_{i, j, n} \otimes \Abf_{2, 1, n} & {\dots} & \Abf_{i, j, n} \otimes \Abf_{2, 2K, n} \\ 
\vdots & \ddots & \vdots \\ 
\Abf_{i, j, n} \otimes \Abf_{2K, 1, n} & {\dots} & \Abf_{i, j, n} \otimes \Abf_{2K, 2K, n} 
\end{bmatrix}.
\end{align*}
Recall that the probability of any client being selected by the server in any iteration $n$ is $\Bar{a} = \Ccal/K$. Consequently, we have 
\begin{align*}
\Ebb \left[ a_{i, n} a_{j, n} \right] = 
\begin{cases} 
\Bar{a} = \frac{\Ccal}{K}                      & i = j  \\
\breve{a} = \frac{\Ccal}{K}\frac{\Ccal-1}{K-1} & i \neq j.
\end{cases}
\end{align*}
Furthermore, we define 
\begin{align*}
\Ncalbf_{il} & = \Nbf_i \otimes \Nbf_{l} \\
\Ncalbf_{0i} & = \Ibf_L \otimes \Nbf_i \\
\Ncalbf_{i0} & = \Nbf_i \otimes \Ibf_L \\
\Ncalbf_i & = \Ncalbf_{0i} + \Ncalbf_{i0}.
\end{align*}
Therefore, we can calculate $\Ebb \left [{\Abf_{i, j, n} \otimes \Abf_{l, m, n} }\right]$ as shown at the top of the following page.
\begin{figure*}[ht!]
\normalsize
\begin{align*}
& \Ebb \left[ { \Abf_{i, j, n} \otimes \Abf_{l, m, n} } \right] = \\
& \begin{cases}
\Ibf_{L^2} - \rho \Bar{a} \Ncalbf_i + \frac{2 \rho \Bar{a}^2 }{\Ccal} \Ncalbf_i + \rho^2 \Bar{a} \Ncalbf_{ii}- \frac{4 \rho^2 \Bar{a}^2 }{\Ccal} \Ncalbf_{ii} & i = j = l = m \in \{1, \ldots, K\} \\
\hspace{5.8mm} + \frac{4 \rho^2 \Bar{a}^2 }{\Ccal^2} \Ncalbf_{ii} &~ \\
\Ibf_{L^2} - \rho \Bar{a} \Ncalbf_{i0} - \rho \Bar{a} \Ncalbf_{0l}+\frac{ 2 \rho \Bar{a}^2 }{\Ccal} \Ncalbf_{i0} + \frac{ 2 \rho \Bar{a}^2 }{\Ccal} \Ncalbf_{0l} & i = j \in \{1, \ldots, K\},~ l = m \in \{1, \ldots, K\},~ i \neq l \\
\hspace{5.8mm} + \rho^2 \Breve{a} \Ncalbf_{il}- \frac{4 \rho^2 \Bar{a} \Breve{a} }{\Ccal} \Ncalbf_{il}+ \frac{4\rho^2 \Breve{a}^2 }{\Ccal^2} \Ncalbf_{il}& \\ 
\frac{2 \rho \Bar{a}^2 }{\Ccal} \Ncalbf_{0i} - \frac{2 \rho^2 \Bar{a}^2 }{\Ccal} \Ncalbf_{ii} + \frac{4 \rho^2 \Bar{a} \Breve{a} }{\Ccal^2} \Ncalbf_{ii} & i = j \in \{1, \ldots, K\},~ l \neq m \in \{1, \ldots, K\},~ i = l,~ j \neq m \\
\frac{2 \rho \Bar{a}^2 }{\Ccal} \Ncalbf_{0l} - \frac{2 \rho^2 \Bar{a} \Breve{a}}{\Ccal} \Ncalbf_{il} + \frac{ 4 \rho^2 \Breve{a} \Bar{a} }{\Ccal^2} \Ncalbf_{il}  & i = j \in \{1, \ldots, K\},~ l \neq m \in \{1, \ldots, K\},~ i \neq l,~ j = m  \\
\frac{2 \rho \Bar{a}^2 }{\Ccal} \Ncalbf_{0l} - \frac{2 \rho^2 \Bar{a} \Breve{a} }{\Ccal} \Ncalbf_{il} + \frac{ 4 \rho^2 \Breve{a}^2 }{\Ccal^2} \Ncalbf_{il} & i = j \in \{1, \ldots, K\},~ l \neq m \in \{1, \ldots, K\},~ i \neq l ,~ j \neq m \\
\frac{- \rho \Bar{a}^2 }{\Ccal} \Ncalbf_{0i} + \frac{ \rho^2 \Bar{a}^2 }{\Ccal} \Ncalbf_{ii} - \frac{ 2 \rho^2 \Bar{a}^3 }{\Ccal^2} \Ncalbf_{ii} & i = j \in \{1, \ldots, K\},~ l \in \{1, \ldots, K\},~ m \in \{K+1, \ldots, 2K\},~ i = l \\
\frac{- \rho \Bar{a}^2 }{\Ccal} \Ncalbf_{0l} + \frac{ \rho^2 \Breve{a} \Bar{a} }{\Ccal} \Ncalbf_{il} - \frac{ 2 \rho^2 \Breve{a} \Bar{a}^2  }{\Ccal^2} \Ncalbf_{il} & i = j \in \{1, \ldots, K\},~ l \in \{1, \ldots, K\},~ m \in \{K+1, \ldots, 2K\},~ i \neq l \\
\Ibf_{L^2} - \rho \Bar{a} \Ncalbf_{i0} + \frac{ 2 \rho \Bar{a}^2 }{\Ccal} \Ncalbf_{i0}  & i = j \in \{1, \ldots, K\},~ m \in \{1, \ldots, K\},~ l = m + K \\
\frac{2 \rho \Bar{a}^2 }{\Ccal} \Ncalbf_{i0} - \frac{2 \rho^2 \Bar{a}^2 }{\Ccal} \Ncalbf_{ii} + \frac{ 4 \rho^2 \Breve{a} \Bar{a} }{\Ccal^2} \Ncalbf_{ii} & i \neq j \in \{1, \ldots, K\},~ l = m \in \{1, \ldots, K\},~ i = l,~ j \neq m \\
\frac{2 \rho \Bar{a}^2 }{\Ccal} \Ncalbf_{i0} - \frac{2 \rho^2 \Breve{a} \Bar{a} }{\Ccal} \Ncalbf_{il} + \frac{ 4 \rho^2 \Breve{a} \Bar{a} }{\Ccal^2} \Ncalbf_{il}  & i \neq j \in \{1, \ldots, K\},~ l = m \in \{1, \ldots, K\},~ i \neq l,~ j = m \\
\frac{2 \rho \Bar{a}^2 }{\Ccal} \Ncalbf_{i0} - \frac{2 \rho^2 \Breve{a} \Bar{a} }{\Ccal} \Ncalbf_{il} + \frac{ 4 \rho^2 \Breve{a}^2 }{\Ccal^2} \Ncalbf_{il} & i \neq j \in \{1, \ldots, K\},~ l = m \in \{1, \ldots, K\},~ i \neq l,~ j \neq m \\
\frac{4 \rho^2 \Bar{a}^2 }{\Ccal^2} \Ncalbf_{ii} & i \neq j \in \{1, \ldots, K\},~ l \neq m \in \{1, \ldots, K\},~ i = l,~ j = m \\
\frac{4 \rho^2 \Bar{a} \Breve{a} }{\Ccal^2} \Ncalbf_{ii} & i \neq j \in \{1, \ldots, K\},~ l \neq m \in \{1, \ldots, K\},~ i = l,~ j \neq m \\
\frac{4 \rho^2 \Bar{a} \Breve{a} }{\Ccal^2} \Ncalbf_{il} & i \neq j \in \{1, \ldots, K\},~ l \neq m \in \{1, \ldots, K\},~ i \neq l,~ j = m \\
\frac{4 \rho^2 \Breve{a}^2 }{\Ccal^2} \Ncalbf_{il} & i \neq j \in \{1, \ldots, K\},~ l \neq m \in \{1, \ldots, K\},~ i \neq l,~ j \neq m \\
\frac{-2 \rho^2 \Bar{a}^3 }{\Ccal^2} \Ncalbf_{ii} & i \neq j \in \{1, \ldots, K\},~ l \in \{1, \ldots, K\},~ m \in \{K+1, \ldots, 2K\},~ i = l \\
\frac{-2 \rho^2 \Breve{a} \Bar{a}^2}{\Ccal^2} \Ncalbf_{il} & i \neq j \in \{1, \ldots, K\},~ l \in \{1, \ldots, K\},~ m \in \{K+1, \ldots, 2K\},~ i \neq l \\
\frac{2 \rho \Bar{a}^2 }{\Ccal} \Ncalbf_{i0} & i \neq j \in \{1, \ldots, K\},~ m \in \{1, \ldots, K\},~ l = m + K \\
\frac{- \rho \Bar{a}^2 }{\Ccal} \Ncalbf_{i0} + \frac{ \rho^2 \Bar{a}^2 }{\Ccal} \Ncalbf_{ii} - \frac{ 2 \rho^2 \Bar{a}^3 }{\Ccal^2} \Ncalbf_{ii} & i \in \{1, \ldots, K\},~ j \in \{K+1, \ldots, 2K\},~ l = m \in \{1, \ldots, K\},~ i = l \\
\frac{-\rho \Bar{a}^2 }{\Ccal} \Ncalbf_{i0} + \frac{ \rho^2 \Breve{a} \Bar{a} }{\Ccal} \Ncalbf_{il} - \frac{ 2 \rho^2 \Breve{a} \Bar{a}^2 }{\Ccal^2} \Ncalbf_{il} & i \in \{1, \ldots, K\},~ j \in \{K+1, \ldots, 2K\},~ l = m \in \{1, \ldots, K\},~ i \neq l \\
\frac{-2 \rho^2 \Bar{a}^3 }{\Ccal^2} \Ncalbf_{ii} & i \in \{1, \ldots, K\},~ j \in \{K+1, \ldots, 2K\},~ l \neq m \in \{1, \ldots, K\},~ i = l  \\
\frac{-2 \rho^2 \Breve{a} \Bar{a}^2 }{\Ccal^2} \Ncalbf_{il} & i \in \{1, \ldots, K\},~ j \in \{K+1, \ldots, 2K\},~ l \neq m \in \{1, \ldots, K\},~ i \neq l \\
 \frac{\rho^2 \Bar{a}^2 }{\Ccal^2} \Ncalbf_{ii} & i=l \in \{1, \ldots, K\},~ j=m \in \{K+1, \ldots, 2K\}\\
\frac{\rho^2 \Breve{a} \Bar{a} }{\Ccal^2} \Ncalbf_{ii} & i=l \in \{1, \ldots, K\},~ j\neq m \in \{K+1, \ldots, 2K\}\\
\frac{\rho^2 \Breve{a} \Bar{a} }{\Ccal^2} \Ncalbf_{il} & i\neq l \in \{1, \ldots, K\},~ j=m \in \{K+1, \ldots, 2K\}\\
\frac{\rho^2 \Breve{a}^2 }{\Ccal^2} \Ncalbf_{il} & i\neq l \in \{1, \ldots, K\},~ j\neq m \in \{K+1, \ldots, 2K\}\\
\frac{- \rho \Bar{a}^2 }{\Ccal} \Ncalbf_{i0} & i \in \{1, \ldots, K\},~ j \in \{K+1, \ldots, 2K\},~ m \in \{1, \ldots, K\},~ l = m + K \\
\Ibf_{L^2} - \rho \Bar{a} \Ncalbf_{0l} + \frac{ 2 \rho \Bar{a}^2 }{\Ccal} \Ncalbf_{0l} & j \in \{1, \ldots, K\},~ i = j + K ,~l = m \in \{1, \ldots, K\} \\
\frac{2 \rho \Bar{a}^2 }{\Ccal} \Ncalbf_{0l} & j \in \{1, \ldots, K\},~ i = j + K ,~ l \neq m \in \{1, \ldots, K\} \\
\frac{- \rho \Bar{a}^2 }{\Ccal} \Ncalbf_{0l} & j \in \{1, \ldots, K\},~ i = j + K ,~ l \in \{1, \ldots, K\},~ m \in \{K+1, \ldots, 2K\}  \\
\Ibf_{L^2} & j \in \{1, \ldots, K\},~ i = j + K ,~ m \in \{1, \ldots, K\},~ l = m + K \\
\mathbf{0}& \text{otherwise.}
\end{cases}
\end{align*}
\hrulefill
\end{figure*}

\section{Evaluation of the spectral radius of $\mathbfcal{Q}$} \label{spectQA}

Recall that 
\begin{equation*}
\Acalbf_n  = 
\left[
\begin{array}{c;{2pt/2pt}c}
    \Acalbf_{n,1} & \Acalbf_{n,2} \\ \hdashline[2pt/2pt]
    \Ibf & \mathbf{0}
\end{array}
\right]
\end{equation*}
and ${\Qcalbf}^\intercal = \Ebb[{\Acalbf}_{n} \otimes_b {\Acalbf}_{n}]$.
In addition, using the definition of the block Kronecker product \cite{tracy1972new}, we can further write ${\Acalbf}_{n} \otimes_b {\Acalbf}_{n}$ as
\begin{align*}
    \small
    \begin{bmatrix}
        \Acalbf_{n,1} \otimes \Acalbf_{n,1} & \Acalbf_{n,1} \otimes \Acalbf_{n,2} & \Acalbf_{n,2} \otimes \Acalbf_{n,1} & \Acalbf_{n,2} \otimes \Acalbf_{n,2} \\
        \Acalbf_{n,1} \otimes \Ibf & \mathbf{0} & \Acalbf_{n,2} \otimes \Ibf & \mathbf{0} \\
        \Ibf \otimes \Acalbf_{n,1} & \Ibf \otimes \Acalbf_{n,2} & \mathbf{0} & \mathbf{0} \\
        \Ibf \otimes \Ibf & \mathbf{0} & \mathbf{0} & \mathbf{0}
    \end{bmatrix}.
\end{align*}
By employing the bilinearity property of the Kronecker product and the fact that $\Acalbf_{n,1} + \Acalbf_{n,2} = \Ibf$, we can sum up each row to an identity matrix. Therefore, ${\Qcalbf}^\intercal$ is a block right-stochastic matrix with a spectral radius of $1$. Hence, the spectral radius of $\Qcalbf$ is also $1$.

\section{Proof of Proposition 1} \label{proof_prop1}

We want to show that 
\begin{align} \label{Ap3_eq1}
    \lim_{n \rightarrow \infty} \boldsymbol{\psi}^\intercal \Qcalbf^n = \lim_{n \rightarrow \infty} \bvec^\intercal \{ \Ebb \left[ {\Acalbf}_{n} \cdots {\Acalbf}_{1} \boldsymbol \psi {\Acalbf}_{1}^{\intercal} \cdots {\Acalbf}_{n}^{\intercal} \right] \} = \mathbf{0}^\intercal,
\end{align}
where $\boldsymbol \Psi = \bvec^{-1} \{ \boldsymbol \psi \}.$
Given that ${\Acalbf}_{n}$ is a right stochastic matrix, we can verify that
\begin{align}
\lim_{n \rightarrow \infty} \prod_{i=1}^n {\Acalbf}_i = \lim_{n \rightarrow \infty} \bar{\Acalbf}^n = 
\left[
\begin{array}{c;{2pt/2pt}c}
    \Bcalbf & \Obf \\ \hdashline[2pt/2pt]
    \Bcalbf & \Obf
\end{array}
\right]
\end{align}
is a constant matrix \cite{anthonisse1977}, where $\Obf$ denotes the zero matrix. This requires 
\begin{align}
    \Bcalbf \Acalbf_{n,1} & = \Acalbf_{n,1} \notag \\
    \Bcalbf \Acalbf_{n,2} & = \Obf, 
\end{align}
which, in light of~\eqref{eq:A_MAT_def}, results in
\begin{align} \label{eq:BNzero}
    \Bcalbf \bar{\Nbf} = \Obf,
\end{align}
where $\bar{\Nbf} = \bdiag \{\Nbf_1,\cdots,\Nbf_K\}$.
Finally, considering \eqref{eq:E_zeta} and \eqref{eq:E_eta}, we can write the expected value in \eqref{Ap3_eq1} as 
\begin{align} \label{Ap3_eq2}
    \left[
\begin{array}{c;{2pt/2pt}c}
    \boldsymbol\Gamma \Bcalbf \bar{\Nbf}^2 \Bcalbf^{\intercal} & \boldsymbol\Gamma \Bcalbf \bar{\Nbf}^2 \Bcalbf^{\intercal} \\ \hdashline[3pt/5pt] 
    \boldsymbol\Gamma \Bcalbf \bar{\Nbf}^2 \Bcalbf^{\intercal} & \boldsymbol\Gamma \Bcalbf \bar{\Nbf}^2 \Bcalbf^{\intercal}
\end{array}
\right] = \Obf,
\end{align}
where the equality is due to \eqref{eq:BNzero} and we have
\begin{align}
\boldsymbol\Gamma = \bar{a} \rho^2 \bdiag \left\{\sigma^2_{\zeta_1} \Ibf, \cdots, \sigma^2_{\zeta_K} \Ibf \right\} + \frac{5 \rho^2}{K^2} \sum_{k=1}^{K} \sigma_{\eta_k}^2 \Ibf.
\end{align}
This concludes the proof.

\bibliographystyle{IEEEtran}
\bibliography{Refs}

\end{document}

%% file: symbols.tex
\newcommand{\sbf}{\mathbf{s}}
\newcommand{\tbf}{\mathbf{t}}
\newcommand{\ubf}{\mathbf{u}}
\newcommand{\vbf}{\mathbf{v}}
\newcommand{\wbf}{\mathbf{w}}

\newcommand{\ybf}{\mathbf{y}}
\newcommand{\zbf}{\mathbf{z}}
\newcommand{\Abf}{\mathbf{A}}

\newcommand{\Ibf}{\mathbf{I}}
\newcommand{\Jbf}{\mathbf{J}}

\newcommand{\Nbf}{\mathbf{N}}
\newcommand{\Obf}{\mathbf{O}}

\newcommand{\Ubf}{\mathbf{U}}

\newcommand{\Xbf}{\mathbf{X}}

\newcommand{\Ccal}{{\mathcal{C}}}
\newcommand{\Dcal}{{\mathcal{D}}}
\newcommand{\Ecal}{{\mathcal{E}}}

\newcommand{\Jcal}{{\mathcal{J}}}

\newcommand{\Lcal}{{\mathcal{L}}}
\newcommand{\Mcal}{{\mathcal{M}}}
\newcommand{\Ncal}{{\mathcal{N}}}

\newcommand{\Scal}{{\mathcal{S}}}

\newcommand{\Ucal}{{\mathcal{U}}}

\newcommand{\Acalbf}{\boldsymbol{\mathcal{A}}}
\newcommand{\Bcalbf}{\boldsymbol{\mathcal{B}}}

\newcommand{\Jcalbf}{\boldsymbol{\mathcal{J}}}

\newcommand{\Ncalbf}{\boldsymbol{\mathcal{N}}}
\newcommand{\Ocalbf}{\boldsymbol{\mathcal{O}}}
\newcommand{\Pcalbf}{\boldsymbol{\mathcal{P}}}
\newcommand{\Qcalbf}{\boldsymbol{\mathcal{Q}}}

\newcommand{\Ucalbf}{\boldsymbol{\mathcal{U}}}

\newcommand{\Wcalbf}{\boldsymbol{\mathcal{W}}}

\newcommand{\Ebb}{\mathbb{E}}
\newcommand{\Rbb}{\mathbb{R}}

\newcommand{\bvec}{{\rm bvec}}

\newcommand{\bdiag}{{\rm bdiag}}
\newcommand{\col}{{\rm col}}
\newcommand{\bcol}{{\rm bcol}}

\newcommand{\tr}{{\rm tr}}
\newcommand{\dB}{{\rm dB}}

\DeclareMathAlphabet\mathbfcal{OMS}{cmsy}{b}{n}

\renewcommand{\IEEEQED}{\IEEEQEDopen}